\documentclass{article}

\usepackage[preprint]{neurips_2026}

\usepackage[utf8]{inputenc} 
\usepackage[T1]{fontenc}    
\usepackage{hyperref}       
\usepackage{url}            
\usepackage{booktabs}       
\usepackage{amsfonts}       
\usepackage{nicefrac}       
\usepackage{microtype}      
\usepackage{xcolor}         

\usepackage{graphicx}
\usepackage{amsmath}
\usepackage{amssymb}
\usepackage{tcolorbox}
\tcbuselibrary{listings}
\usepackage{subcaption}
\usepackage{algorithm}
\usepackage{tipa}
\usepackage{algpseudocode}

\title{\methodname{}}

\author{%
  {Atharva Naik$^\dagger$ \quad Yash Mathur$^\clubsuit$ \quad Prakam$^\clubsuit$ \quad Carolyn Rose$^\dagger$ \quad David Mortensen$^\dagger$} \\
  $^\dagger$Carnegie Mellon University, $^\clubsuit$Independent Researcher  \\
  \texttt{\{arnaik,cprose,dmortens\}@cs.cmu.edu} \\ \texttt{prakam@iitdalumni.com}  \\
}
\newcommand{\methodname}{\textsc{ReaComp}}
\title{\methodname{}: Compiling LLM Reasoning into Symbolic Solvers for Efficient Program Synthesis}

\begin{document}

\maketitle

\begin{abstract}
LLMs can solve program synthesis tasks but remain inefficient and unreliable on hard instances requiring large combinatorial search.
Given a small set of reasoning traces, we use coding agents to compile them into reusable symbolic program synthesizers over constrained DSLs.
The resulting solvers require no LLM calls at test time and are strong standalone systems: symbolic solver ensembles reach 91.3\% accuracy on PBEBench-Lite and 84.7\% on PBEBench-Hard, outperforming LLMs with test-time scaling for the latter by +16.3 percentage points at zero LLM inference cost. 
They also complement LLM search, improving PBEBench-Hard accuracy from 68.4\% to 85.8\% while reducing reported token usage by 78\%, and raising SLR-Bench hard-tier accuracy from 34.4\% to 58.0\% in a neuro-symbolic hybrid setting.
Compared to directly using coding agents as per-instance solvers, induced solvers are substantially more Pareto-efficient, amortizing a small one-time construction cost over many zero-token executions. 
Finally, most solvers transfer zero-shot to a real historical linguistics task — predicting sound changes in natural language data — reaching 80.1\% accuracy under ensembling and recovering some plausible linguistic rules. 
Together, these results show that reasoning traces can be compiled into reusable symbolic solvers that solve many tasks directly, complement LLM inference on hard cases, and provide a scalable route to domain-general solver induction.
We release code and data for reproducibility\footnote{\url{https://github.com/cmu-llab/ReaComp}}
\end{abstract}

\begin{figure*}[!tbh]
    \centering
    \includegraphics[width=\textwidth]{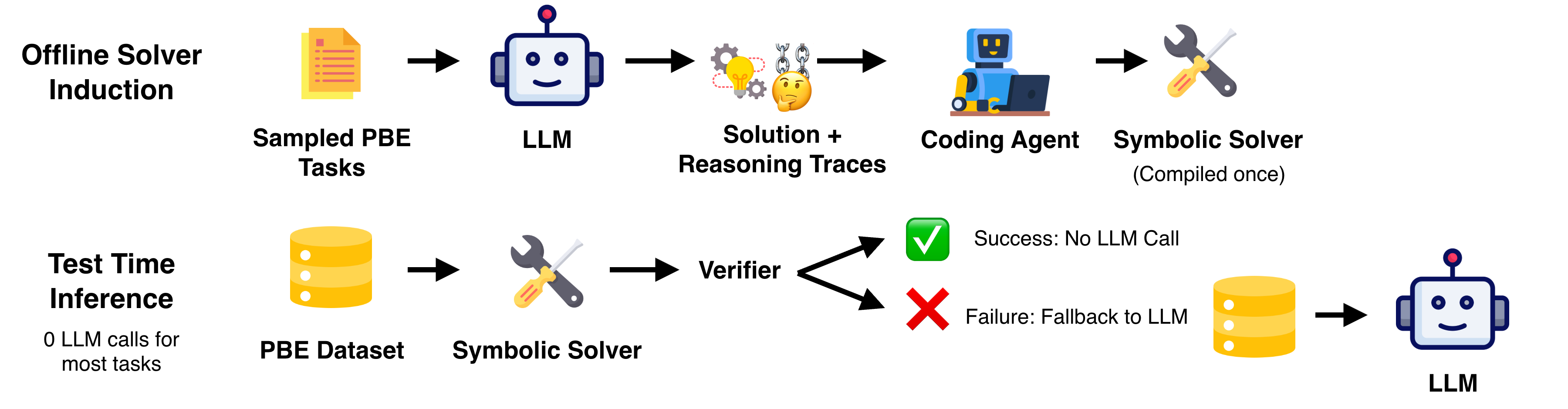}
    \caption{\methodname{}: \textbf{offline solver induction} compiles LLM reasoning traces into a reusable symbolic solver; \textbf{test-time inference} applies the solver directly (zero LLM cost) and falls back to LLM search only for unresolved cases.}
    \label{fig:method}
\end{figure*}

\vspace{-8pt}
\section{Introduction}
\label{sec:intro}
\vspace{-8pt}
The Programming by Example (PBE) paradigm \citep{Gulwani2011} provides a controlled setting for studying structured reasoning and compositional generalization, with applications spanning code transformation, data wrangling, text processing, and historical linguistics. 
Recent work shows that large language models (LLMs), when combined with scaling strategies such as Best-of-$K$ sampling and iterative refinement, can achieve strong performance on PBE tasks \citep{li2024programming, naik2025pbebench}. 
In particular, prior work like PBEBench \citep{naik2025pbebench} demonstrates that sufficiently powerful reasoning models, together with these scaling strategies, can achieve high accuracy on PBE tasks without explicit symbolic structure.

However, this performance comes at a substantial computational cost. 
Through analysis of reasoning traces and scaling behavior, we find that harder instances require disproportionately more attempts and token usage (Figure~\ref{fig:pbebench_lite_by_cl}), often due to failures of compositional generalization and inefficient search. 
Even strong reasoning models such as \texttt{gpt-oss-120b} exhibit unstable behaviors, including looping within a single attempt and redundant exploration across attempts in iterative refinement settings. 
Moreover, we observe that LLM-based search performs well on short, simple programs but degrades significantly on longer, compositional transformations. 
These patterns suggest that, despite their capability, LLMs lack an explicit mechanism for structuring search over candidate programs, leading to wasted computation on difficult tasks, especially as solution length increases.

A natural approach is to augment LLMs with tools or learned libraries, including approaches that induce reusable abstractions from experience \citep{wang2024trove, stengel2024regal, yue2025toollibgen}. 
While these methods demonstrate that LLMs can acquire useful structure, they typically rely on iterative reasoning and tool invocation at inference time, and thus remain sensitive to compute budgets. 
Compute-matched evaluations show that their gains are largely driven by increased inference-time compute rather than effective reuse of learned tools, with limited evidence of consistent tool reuse across instances \citep{sesterhenn2025compute}. 
As a result, they continue to perform a largely unstructured search over the solution space.

In this work, we propose \methodname{}: instead of improving inference-time reasoning, we \emph{compile} it into a reusable symbolic structure. 
Our approach consists of two stages: \textbf{offline solver induction} and \textbf{test-time inference}. 
In the offline phase, we induce \emph{symbolic solvers} directly from LLM reasoning traces by using coding agents to distill recurring strategies and failure modes into standalone program synthesizers over a constrained DSL. 
These solvers require no LLM calls at test time and can solve a large fraction of tasks directly, often matching or exceeding LLM search on harder instances.
At test time, the solver provides a cheap, structured first pass, and the LLM is invoked only as a fallback on unresolved cases \citep{zhang2025pbe}.

We evaluate \methodname{} on recent, challenging benchmarks — PBEBench \citep{naik2025pbebench} and SLR-Bench \citep{helff2025slr} — and show that induced symbolic solvers achieve strong standalone performance while significantly improving inference efficiency.
Symbolic solver ensembles reach 91.3\% accuracy on PBEBench-Lite and 84.7\% on PBEBench-Hard, outperforming a test-time scaling LLM baseline on Hard by +16.3 percentage points (pp) at zero LLM inference cost.
On SLR-Bench, symbolic solvers match frontier LLMs on the hardest tier and, when combined with direct feedback, improve hard-tier accuracy from 34.4\% to 58.0\%.
Hybrid inference reduces token usage by up to 78\%. By amortizing a small one-time construction cost over many zero-token executions, induced solvers Pareto-dominate approaches that directly apply the same coding agent to each task individually.
Finally, most solvers transfer zero-shot to forward reconstruction in historical linguistics — predicting how sounds change across related languages — reaching 80.1\% accuracy under ensembling and recovering plausible sound change rules.

\textbf{Contributions.}
We propose \methodname{}, contributing:
(1) a method for inducing reusable symbolic solvers from LLM reasoning traces via coding agents;
(2) evidence that induced solvers are strong standalone systems, outperforming LLM test-time scaling on hard instances at zero per-task LLM inference cost;
(3) a hybrid neuro-symbolic pipeline that Pareto-dominates per-task agentic effort, reducing token usage by up to 78\% while achieving state-of-the-art results;
(4) a demonstration that most induced solvers generalize zero-shot across distributions, with a case study on forward reconstruction in historical linguistics.
\section{Method}
\label{sec:method}
\label{section:background}

\paragraph{Problem formulation.}
Let $\mathcal{X}$ and $\mathcal{Y}$ denote input and output spaces.
A PBE task is a finite set of examples $\mathcal{D} = \{(x_i, y_i)\}_{i=1}^n$.
The goal is to infer a program $p \in \mathcal{P}$ (from a DSL $\mathcal{P}$) such that $p(x_i)=y_i$ for all examples.
We use a task-level reward $R(p;\mathcal{D}) \in [0,1]$, with $R=1$ indicating a fully correct solution.
We propose \methodname{}, a two-stage approach that (1) compiles LLM reasoning traces into symbolic solvers and (2) uses these solvers in a hybrid inference pipeline with LLM-based search.

\subsection{LLM Search Procedures}

An LLM defines a conditional distribution $p \sim \pi_\theta(\cdot \mid \mathcal{D})$.
We consider two standard test-time scaling strategies, both selecting $p^* = \arg\max_{p \in \mathcal{C}} R(p;\mathcal{D})$:

\textbf{Best-of-$K$ (BoK).}
Sample $K$ programs independently and select the highest-scoring candidate: $\mathcal{C} = \{p_k\}_{k=1}^K$. \\
\textbf{Direct Feedback (DF).}
Generate a trajectory $\{p^{(t)}\}_{t=1}^T$ via iterative refinement conditioned on verifier feedback: $\mathcal{C} = \{p^{(t)}\}_{t=1}^T$.
Early stopping is applied for both when $R=1$.

Both define candidate sets $\mathcal{C}$ used in our hybrid inference.

\subsection{Offline Solver Induction}

Solver induction is a one-time offline procedure that compiles reasoning traces into a standalone symbolic program synthesizer.
Given training tasks $\{\mathcal{D}_j\}_{j=1}^m$ and corresponding traces $\{\tau_j\}_{j=1}^m$, we induce a solver $S_\phi : \mathcal{D} \mapsto \mathcal{P}$ by approximately maximizing $\sum_j R(S_\phi(\mathcal{D}_j); \mathcal{D}_j)$ via a coding agent that distills reusable structure and failure modes from traces.

\textbf{Trace dataset.}
We construct a dataset $\mathcal{T} = \{(\mathcal{D}_j, \tau_j)\}_{j=1}^m$ of reasoning traces generated by an LLM (e.g., \texttt{gpt-oss-120b}), balanced across task difficulty and success/failure. Each trace contains intermediate reasoning steps, candidate programs, and outcomes. The induction tasks are a small subset of the benchmark tasks; no ground-truth programs are accessed during induction. The coding agent observes only the input-output examples available to any solver at test time.

\textbf{Coding-agent-based induction.}
We use a coding agent $\mathcal{A}_\psi$ to iteratively construct a solver implementation. The agent operates over a sandboxed container containing the trace dataset and the verifier code (read-only access to avoid hacking the verifier). 
Access to the verifier allows the agent to test candidate solvers and iteratively debug or refine them.
At each step, the agent proposes edits to the workspace, executes code, and refines the solver based on performance over $\mathcal{T}$. The final output is a standalone Python solver file \texttt{SOLVER.py} and a corresponding algorithm description file (\texttt{SOLVER\_ALGORITHM.md}).

We experiment with two coding agents for solver induction:
(i) \textbf{Claude Code (CC)} using \texttt{claude-sonnet-4-6} in an unconstrained interactive session, and 
(ii) \textbf{Qwen + OpenHands (QO)} using \texttt{Qwen3.6-35B-A3B} with a bounded trajectory (max 500 steps per run) to model a resource-constrained setting.
For both agents, we construct a trace dataset of approximately 100 examples per benchmark (PBEBench and SLR-Bench), sampled to balance task difficulty and outcome (success vs.\ failure), enabling the agent to learn both effective strategies and common failure modes.
We additionally perform ablations with QO on PBEBench to study the impact of reasoning traces (CoT vs.\ no-CoT), trace dataset size (100, 48, 12), and solver algorithm variance across three independent induction trajectories with the same trace dataset.

\subsection{Efficient Hybrid Inference}
At test time, the solver proposes a set of candidates $\mathcal{C}_S = S_\phi(\mathcal{D})$ and selects the best $p_S = \arg\max_{p \in \mathcal{C}_S} R(p;\mathcal{D})$; if $R(p_S;\mathcal{D})=1$ it is accepted at zero LLM cost.
Otherwise we fall back to LLM search (BoK or DF) and return $\arg\max_{p \in \mathcal{C}_L} R(p;\mathcal{D})$, where $\mathcal{C}_L$ is the LLM candidate set.
This amortizes solver construction over many zero-token executions: $S_\phi$ handles the majority of tasks directly while LLM search is reserved for harder residual cases.

Building prompts, solver interfaces, and agent configurations are provided in Appendix~\ref{app:prompts}. We also formalize both the described steps in algorithms~\ref{alg:solver_induction} and \ref{alg:hybrid}. $w_t$ denotes the evolving code sandbox, $a_t$ is a coding-agent action (e.g., edit, execution, or file write), and $\mathcal{A}_\psi$ is the agent policy over actions. 
$\textsc{ExtractSolver}$ parses the workspace into an executable solver $S_{\phi_t}$ when a valid implementation (e.g., \texttt{SOLVER.py}) is present.
$q_t$ measures solver quality on the trace dataset via the verifier, effectively capturing the fit of the solver over the training trace dataset.

\begin{algorithm}[t]
\caption{Offline Solver Induction via Coding Agent}
\label{alg:solver_induction}
\begin{algorithmic}[1]
\Require trace LLM $\pi_\theta$, verifier $R$, tasks $\{\mathcal{D}_j\}_{j=1}^m$, coding agent $\mathcal{A}_\psi$, step budget $N$
\State $\tau_j \leftarrow \textsc{Trace}(\pi_\theta, R, \mathcal{D}_j)$ for $j=1,\ldots,m$
\State $\mathcal{T} \leftarrow \{(\mathcal{D}_j, \tau_j)\}_{j=1}^m$
\State Initialize workspace $w_0$ (empty codebase)
\For{$t = 0, \ldots, N-1$}
    \State $a_t \sim \mathcal{A}_\psi(\cdot \mid w_t, \mathcal{T}, R)$
    \State $w_{t+1} \leftarrow \textsc{Apply}(w_t, a_t)$
    \State $S_{\phi_t} \leftarrow \textsc{ExtractSolver}(w_{t+1})$
    \If{$S_{\phi_t}$ is valid}
        \State $q_t \leftarrow \sum_{j=1}^m R(S_{\phi_t}(\mathcal{D}_j);\mathcal{D}_j)$
    \EndIf
    \If{$\textsc{Done}(a_t)$}
        \State \textbf{break}
    \EndIf
\EndFor
\State $t^* \leftarrow \arg\max_t q_t$
\State \Return solver $S_\phi \leftarrow S_{\phi_{t^*}}$
\end{algorithmic}
\end{algorithm}

In practice, $t^*$ corresponds to the agent's final committed solver state at termination; the agent self-evaluates on $\mathcal{T}$ inline during the run to guide refinement rather than through post-hoc checkpoint selection across all steps.

\begin{algorithm}[t]
\caption{Hybrid Test-Time Inference}
\label{alg:hybrid}
\begin{algorithmic}[1]
\Require task $\mathcal{D}$, solver $S_\phi$, verifier $R$, LLM policy $\pi_\theta$, mode $M$
\State $\mathcal{C}_S \leftarrow S_\phi(\mathcal{D})$ \Comment{may return multiple candidates}
\State $p_S \leftarrow \arg\max_{p \in \mathcal{C}_S} R(p;\mathcal{D})$
\If{$R(p_S;\mathcal{D}) = 1$}
    \State \Return $p_S$
\EndIf
\State $\mathcal{C}_L \leftarrow \textsc{LLMSearch}(\pi_\theta, R, \mathcal{D}, M)$
\State \Return $\arg\max_{p \in \mathcal{C}_L} R(p;\mathcal{D})$
\end{algorithmic}
\end{algorithm}
\section{Experimental Setup}
\label{sec:experimental_setup}

We evaluate whether LLM reasoning traces can be compiled into reusable symbolic solvers that improve accuracy, efficiency, and performance on hard compositional program induction tasks. Experiments span two domains: programming-by-example (PBE) string transformation and scalable logical reasoning (SLR).

\textbf{Benchmarks and metrics.}
We evaluate on PBEBench \cite{naik2025pbebench} (PBEBench-Lite and PBEBench-Hard) and SLR-Bench \cite{helff2025slr}. 
PBEBench requires synthesizing ordered cascades of string rewrite programs from input-output examples. PBEBench-Lite contains 1,008 tasks with cascade lengths 2--5 (approximately 252 tasks per level) and a maximum program budget of 5. PBEBench-Hard contains 1,216 tasks with cascade lengths 2--20 (64 tasks per level) and a maximum program budget of 20. 
SLR-Bench contains 1,000 inductive logic programming tasks across curriculum levels 1--20, grouped into four tiers (basic, easy, medium, hard). Each task requires inducing a Prolog rule of the form \texttt{eastbound(T) :- Body.} from background facts and labeled examples.
For PBEBench, programs are \texttt{replace(A,B)} sequences with $|A|\in[1,3]$, $|B|\in[0,3]$, evaluated by exact execution; for SLR-Bench, solutions are Prolog rules scored by the official \texttt{partial\_score} reward. Evaluation is fully deterministic and verifier-based: a task is solved only when the reward is 1.0.
We report accuracy (perfect solution found), mean reward, token usage and cost (token accounting discussed in \S\ref{sec:appendix:experimental_details:token_accounting}). 
We additionally report a complexity gap $\Delta$ between predicted and ground-truth programs. 
For PBEBench, we follow the complexity measure defined in the original paper, but compute it \emph{only over correctly solved tasks}: incorrect outputs are often degenerate and would bias the mean downward if included.
For SLR-Bench, $\Delta$ is defined analogously using rule complexity (number of top-level body literals excluding \texttt{has\_car/2}), and is also reported only on correct solutions. 
We further report edit similarity on PBEBench and tier-wise accuracy on SLR-Bench.
\\ 
\\
\textbf{LLM baselines.}
We instantiate the LLM search procedures described in \S\ref{sec:method} using \texttt{gpt-oss-120b} served via vLLM.
BoK (parallel sampling) and DF (sequential refinement with verifier feedback) together cover the two dominant paradigms for test-time LLM scaling.
We evaluate BoK with $K=32$ and DF with up to 32 refinement steps and early stopping; temperature is fixed at 1.0 for both.
Token budgets are 32,768 tokens per attempt on PBEBench-Lite and SLR-Bench, and 16,384 on PBEBench-Hard.
On PBEBench-Hard, we evaluate BoK only, as DF becomes prohibitively expensive and prone to lock-in for long cascades.
We additionally include a coding-agent baseline (\texttt{Qwen3.6-35B-A3B} via OpenHands, 100-step budget per task) that directly solves each task in isolation, providing a comparison point that isolates the contribution of solver induction over per-task agentic effort.
\\
\\
\textbf{Symbolic solver induction and hybrid inference.}
We induce solvers offline as described in \S\ref{sec:method}, using \textbf{CC} (\texttt{claude-sonnet-4-6}) and \textbf{QO} (\texttt{Qwen3.6-35B-A3B}). 
Because coding-agent stochasticity yields distinct implementations, we report individual solvers and unions across runs: CC+QO denotes the union of the CC and QO solvers, while \textbf{``All Symbolic''} denotes all 6 QO solvers from the PBEBench ablations plus the CC solver.
We omit \textbf{``All Symbolic''} for SLR-Bench, where only one QO and one CC solver are induced.
All solvers are standalone, require no LLM calls at inference time, and are effectively deterministic (${<}1\%$ per-task flips on PBEBench-Hard).
We evaluate the hybrid procedure from \S\ref{sec:method}, applying symbolic solvers before LLM search.

\begin{table*}[!tbh]
\centering
\caption{\textbf{PBEBench-Lite results.} We compare reported single-attempt baselines, direct coding agent solving, test-time scaling, coding agent-induced symbolic solvers, and efficiency-oriented hybrids.
$\Delta$ denotes predicted minus ground-truth cascade complexity on correct solutions only.}
\smallskip
\resizebox{\textwidth}{!}{%
\begin{tabular}{lcccccc}
\toprule
\textbf{System} & \textbf{Acc\%} & \textbf{Mean Reward} & \textbf{Edit Sim} & $\boldsymbol{\Delta}$ & \textbf{Tokens (M)} & \textbf{Cost (\$)} \\
\midrule
\multicolumn{7}{l}{\textit{LLM-only methods}} \\
gpt-oss-120b (single)$^\dagger$ & 62.5 & --- & 69.9 & --- & --- & --- \\
GPT-5 (single)$^\dagger$ & 72.4 & --- & 76.5 & --- & --- & --- \\
Qwen3.6-35B-A3B (OpenHands) & 87.2 & 0.9544 & 94.9 & +1.47 & 395.3 & 85.38 \\
DF-32 (gpt-oss-120b) & 92.4 & 0.9796 & 97.3 & +2.11 & 111.1 & 16.74 \\
BoK (gpt-oss-120b) & 93.8 & 0.9808 & 97.8 & +2.19 & 68.0 & 12.20 \\
\midrule
\multicolumn{7}{l}{\textit{Symbolic solvers}} \\
CC & 80.4 & 0.9438 & 93.7 & +3.00 & 0 & 2.00 \\
QO & 65.7 & 0.9022 & 87.4 & +3.01 & 0 & 0.85 \\
CC + QO (union) & 84.6 & 0.9607 & 94.9 & +2.16 & 0 & 2.85 \\
All Symbolic & 91.3 & 0.9772 & 96.6 & +1.88 & 0 & 7.11 \\
\midrule
\multicolumn{7}{l}{\textit{Hybrid (symbolic + LLM fallback)}} \\
DF + QO & 92.9 & 0.9810 & 97.5 & +2.89 & 90.2 & 14.35 \\
DF + CC & 93.1 & 0.9815 & 97.6 & +3.00 & 80.5 & 13.97 \\
DF + All Symbolic & 93.2 & 0.9817 & 97.6 & +2.18 & 78.7 & 18.80 \\
BoK + QO & 93.8 & 0.9808 & 97.8 & +2.94 & 50.7 & 9.97 \\
BoK + CC & 93.9 & 0.9810 & 97.8 & +2.94 & 45.6 & 10.22 \\
BoK + All Symbolic & \textbf{93.9} & \textbf{0.9810} & \textbf{97.8} & +2.19 & \textbf{43.5} & 14.94 \\
\bottomrule
\end{tabular}
}
\label{tab:pbebench_lite}
\end{table*}

\begin{figure*}[t]
\centering
\begin{subfigure}{0.48\textwidth}
    \centering
    \includegraphics[width=\linewidth]{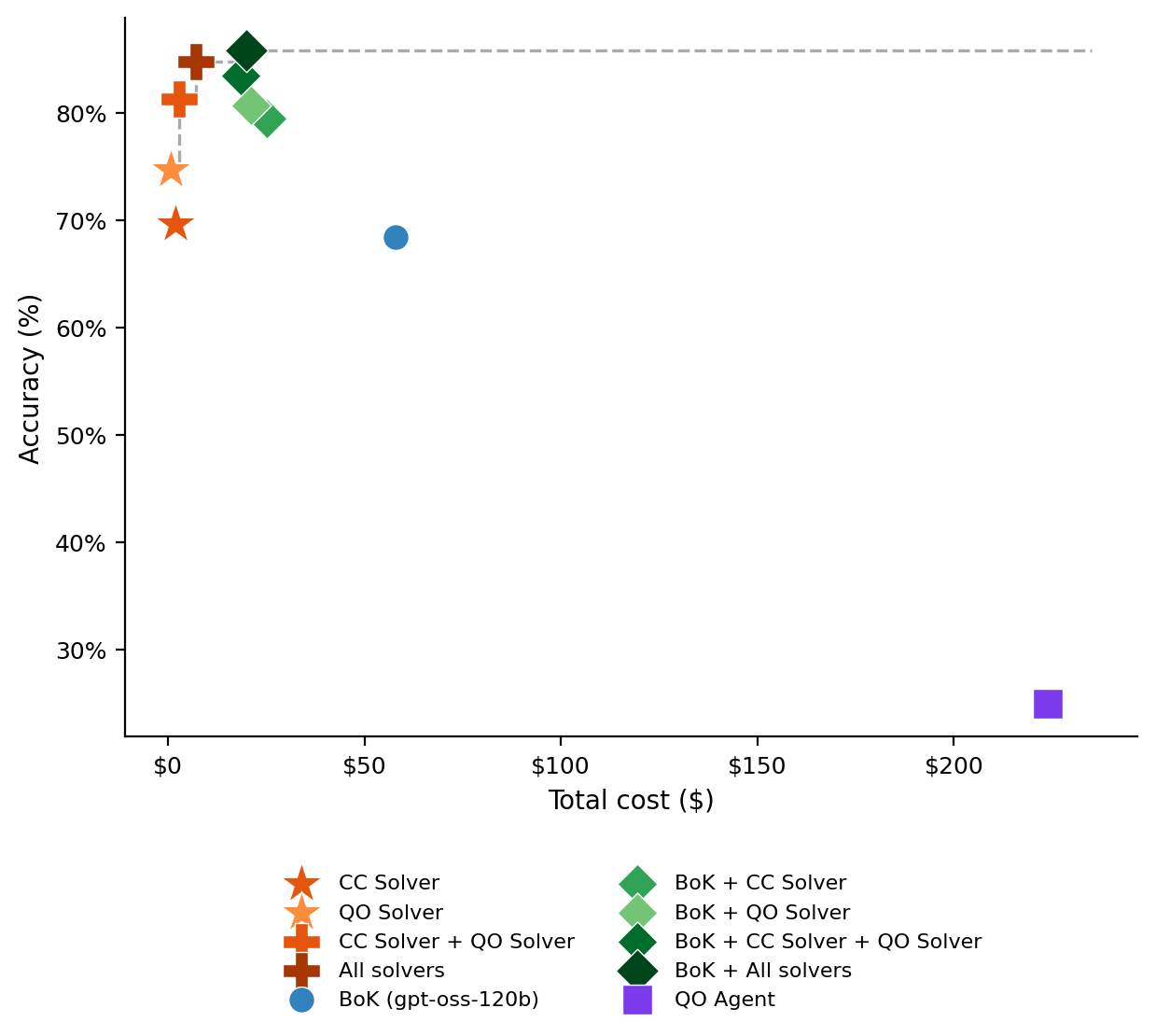}
    \caption{PBEBench-Hard. Frontier: QO $\rightarrow$ CC+QO $\rightarrow$ All solvers $\rightarrow$ BoK+All solvers.}
    \label{fig:pareto_pbe_hard}
\end{subfigure}\hfill
\begin{subfigure}{0.48\textwidth}
    \centering
    \includegraphics[width=\linewidth]{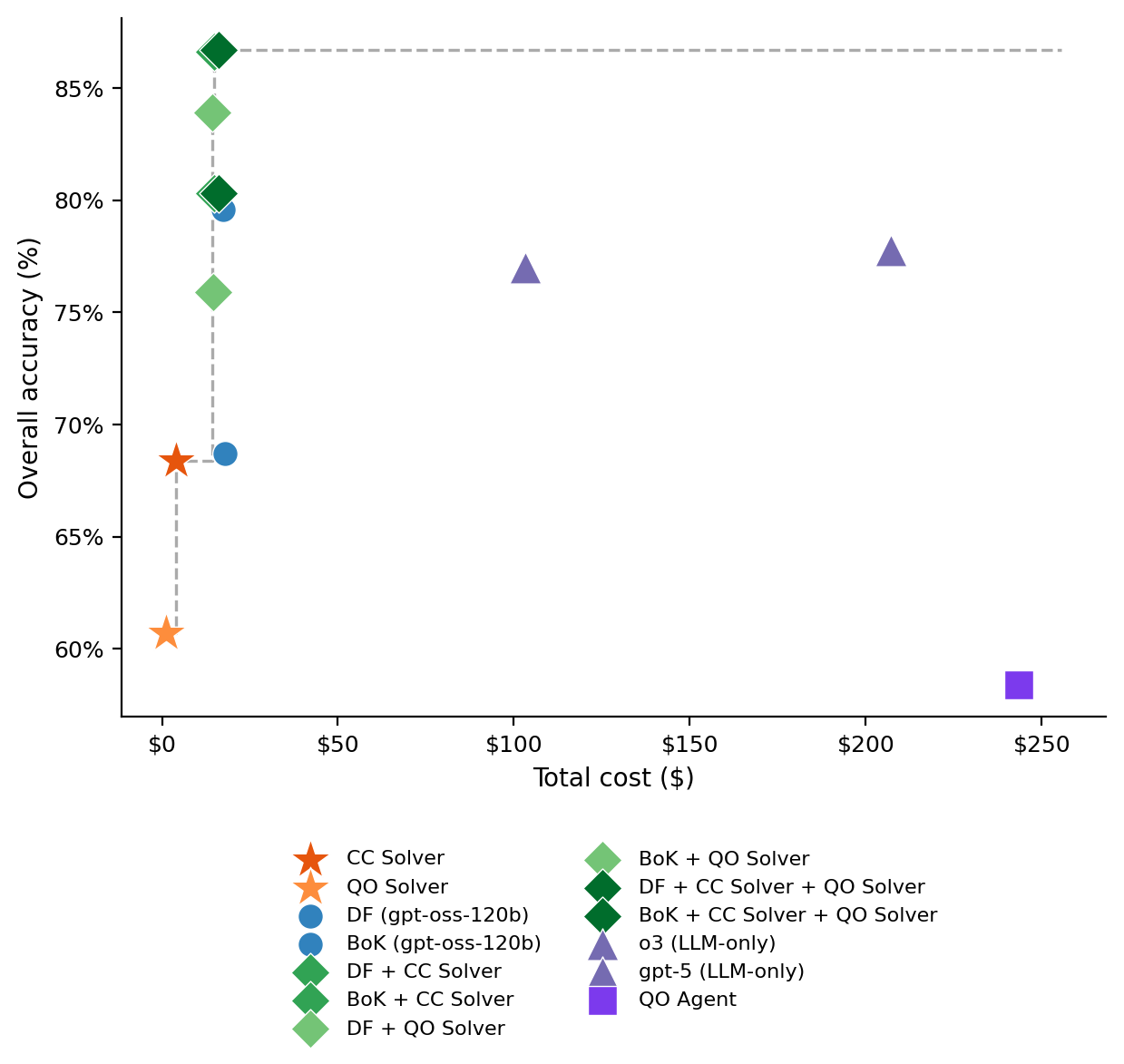}
    \caption{SLR-Bench. Frontier: QO $\rightarrow$ CC $\rightarrow$ DF+QO $\rightarrow$ DF+CC $\rightarrow$ DF+CC+QO.}
    \label{fig:pareto_slr}
\end{subfigure}
\caption{\textbf{Cost--performance Pareto frontiers.} Symbolic solvers dominate the low-cost regime, while hybrid methods achieve the best trade-offs and absolute performance.}
\label{fig:pareto_frontiers}
\end{figure*}

\begin{table*}[!tbh]
\centering
\caption{\textbf{PBEBench-Hard results.} We compare direct coding-agent solving, LLM test-time search, coding-agent-induced symbolic solvers, and efficiency-oriented hybrids on long-horizon cascade synthesis.
$\Delta$ denotes predicted minus ground-truth cascade complexity on correct solutions only.}
\smallskip
\resizebox{\textwidth}{!}{%
\begin{tabular}{lcccccc}
\toprule
\textbf{System} & \textbf{Acc\%} & \textbf{Mean Reward} & \textbf{Edit Sim} & $\boldsymbol{\Delta}$ & \textbf{Tokens (M)} & \textbf{Cost (\$)} \\
\midrule
\multicolumn{7}{l}{\textit{LLM-only methods}} \\
BoK & 68.4 & 0.9428 & 89.9 & +5.14 & 332.1 & 57.83 \\
Qwen3.6-35B-A3B (OpenHands) & 24.9 & 0.8411 & 73.2 & +1.36 & 1046.8 & $\sim$224 \\
\midrule
\multicolumn{7}{l}{\textit{Symbolic solvers}} \\
CC Solver & 69.7 & 0.9873 & 97.2 & +8.06 & 0 & 2.00 \\
QO Solver & 74.7 & 0.9836 & 96.8 & +5.26 & 0 & 0.85 \\
CC + QO & 81.2 & 0.9905 & 98.3 & +5.35 & 0 & 2.85 \\
All Symbolic & 84.7 & 0.9920 & 98.6 & +4.56 & 0 & 7.11 \\
\midrule
\multicolumn{7}{l}{\textit{Hybrid (symbolic + LLM fallback)}} \\
BoK + CC & 79.4 & 0.9508 & 91.4 & +7.82 & 130.0 & 25.08 \\
BoK + QO & 80.7 & 0.9496 & 91.3 & +5.48 & 114.5 & 21.26 \\
BoK + CC + QO & 83.5 & 0.9531 & 91.9 & +5.48 & 87.6 & 18.48 \\
BoK + All Symbolic & \textbf{85.8} & \textbf{0.9570} & \textbf{92.7} & +4.64 & \textbf{71.6} & \textbf{19.89} \\
\bottomrule
\end{tabular}%
}
\label{tab:pbebench_hard}
\end{table*}

\begin{table*}[!tbh]
\centering
\caption{\textbf{SLR-Bench results.} We compare reported frontier-model baselines \cite{helff2025slr}, direct coding-agent solving, LLM test-time search, coding-agent-induced symbolic solvers, and efficiency-oriented hybrids on inductive Prolog rule synthesis.
Tier columns report accuracy by curriculum tier (Basic / Easy / Medium / Hard), each with 250 tasks.
$\Delta$ denotes predicted minus ground-truth rule complexity on correct solutions only.
}
\smallskip
\resizebox{\textwidth}{!}{%
\begin{tabular}{lcccccccc}
\toprule
\textbf{System} & \textbf{Acc\%} & \textbf{Basic} & \textbf{Easy} & \textbf{Medium} & \textbf{Hard} & \textbf{Tokens (M)} & \textbf{Cost (\$)} & $\boldsymbol{\Delta}$ \\
\midrule
\multicolumn{9}{l}{\textit{LLM-only methods}} \\
o3$^\dagger$ & 77.8 & 99 & 93 & 74 & 45 & 4.30 & 207.24 & --- \\
gpt-5$^\dagger$ & 77.0 & 100 & 90 & 72 & 46 & 16.40 & 103.13 & --- \\
BoK & 68.7 & 100 & 100 & 57.6 & 17.2 & 225.3 & 17.88 & -0.611 \\
DF-32 & 79.6 & 100 & 99.6 & 84.4 & 34.4 & 224.2 & 17.43 & -0.834 \\
Qwen3.6-35B-A3B (OpenHands) & 58.4 & 100 & 86.0 & 32.8 & 14.8 & 1232.6 & 243.41 & -0.354 \\
\midrule
\multicolumn{9}{l}{\textit{Symbolic solvers}} \\
CC Solver & 68.4 & 100 & 78.4 & 48.4 & 46.8 & 0 & 4.01 & -0.756 \\
QO Solver & 60.7 & 100 & 71.2 & 34.4 & 37.2 & 0 & 1.28 & -1.087 \\
\midrule
\multicolumn{9}{l}{\textit{Hybrid (symbolic + LLM fallback)}} \\
BoK + QO & 75.9 & 100 & 100 & 62.8 & 40.8 & 162.8 & 14.49 & -1.063 \\
BoK + CC & 80.3 & 100 & 100 & 68.4 & 52.8 & 132.4 & 14.81 & -0.796 \\
BoK + CC + QO & 80.3 & 100 & 100 & 68.4 & 52.8 & 132.4 & 16.09 & -0.795 \\
\midrule
DF + QO & 83.9 & 100 & 99.6 & 86.4 & 49.6 & 166.1 & 14.29 & -1.123 \\
DF + CC & 86.6 & 100 & 99.6 & 88.8 & 58.0 & 138.8 & 14.91 & -0.924 \\
DF + CC + QO & 86.7 & 100 & 99.6 & 88.8 & 58.4 & 138.8 & 16.19 & -0.924 \\
\bottomrule
\end{tabular}%
}
\label{tab:slr_bench}
\end{table*}

\vspace{-15pt}
\section{Results}
\vspace{-5pt}

For symbolic solvers, we report one-time solver construction cost (no per-task LLM usage).
For hybrid methods, reported tokens reflect \emph{LLM fallback inference only} (tasks solved by the symbolic solver incur zero tokens); reported cost additionally includes the \emph{one-time solver construction cost}.
$\dagger$ denotes reported results from prior work.

\textbf{PBEBench-Lite.}
Results are shown in Table~\ref{tab:pbebench_lite} (full details in Appendix~\ref{app:lite}). 
The QO solver shown corresponds to ``100 examples + CoT, run 2'' out of the 6 ablations in Appendix~\ref{app:solver_ablations}.
LLM baselines achieve strong performance with test-time scaling (BoK: 93.8\%, DF: 92.4\%), but incur substantial token usage. 
A direct coding-agent baseline (OpenHands) achieves 87.2\% accuracy, outperforming individual symbolic solvers but falling below LLM scaling methods at substantially higher cost. 
Induced symbolic solvers are competitive as standalone systems, with the CC solver reaching 80.4\% accuracy at zero inference cost, outperforming single-attempt LLM baselines. 
Combining multiple solvers yields 91.3\% accuracy without any LLM usage, within 2.5pp of BoK. 
Hybrid inference substantially reduces LLM token usage: BoK + All Symbolic lowers total tokens from 68.0M to 43.5M (-36\%) while maintaining peak accuracy (93.9\%). 
Among hybrids, BoK + QO achieves Pareto efficiency, matching peak BoK accuracy while reducing both token usage and total cost (\$9.97 vs.\ \$12.20).
BoK + CC yields a marginal additional accuracy gain of 0.1pp at comparable cost (\$10.22), as the one-time solver construction overhead (\$2.00 CC, \$0.85 QO) is negligible relative to per-task inference savings.
Across all methods, predicted cascade complexity remains close to ground truth ($\Delta \approx +2$), indicating that accuracy gains are not attributable to trivial or degenerate programs.

\textbf{PBEBench-Hard.}
Results are shown in Table~\ref{tab:pbebench_hard} (full details in Appendix~\ref{app:hard}).
On long-horizon cascade synthesis (lengths 2--20), the performance regime shifts substantially relative to Lite: the direct coding-agent baseline (OpenHands) achieves only 24.9\% accuracy at high token cost (\$224), while induced symbolic solvers surpass LLM BoK scaling (68.4\%) at zero per-task LLM cost. 
QO solver (``100 examples + CoT, run 2'') reaches 74.7\% and CC solver reaches 69.7\%.
The All Symbolic union (6 QO solvers and CC solver) further improves to 84.7\%, exceeding BoK by 16.3pp with no LLM inference.
Hybrid inference provides the largest absolute gains on this split: BoK + All Symbolic reaches 85.8\% while reducing reported token cost by 78\% relative to BoK alone (71.6M vs.\ 332.1M tokens), as the solver eliminates LLM cost on the 84.7\% of tasks it handles correctly.
Predicted cascade complexity substantially overshoots ground truth ($\Delta \approx +5$--$+8$) on Hard, reflecting the difficulty of recovering compact solutions at long cascade lengths; the hybrid partially mitigates this by selecting simpler BoK candidates for tasks where the solver fails.

\textbf{SLR-Bench.}
Results are shown in Table~\ref{tab:slr_bench} (full details in Appendix~\ref{app:slr}).
Performance across tiers is highly non-uniform: BoK achieves 100\% on Basic and Easy but collapses to 17.2\% on Hard, while DF is more balanced (34.4\% Hard) at comparable token cost.
The direct coding-agent baseline consumes the most tokens by far while achieving the lowest overall accuracy and the weakest Hard-tier performance of any method we evaluate.
Induced symbolic solvers exhibit the complementary profile: the CC Solver reaches 46.8\% on Hard — matching the best reported frontier-model results (o3: 45\%, GPT-5: 46\%) at zero per-task LLM cost.
Hybrid inference leverages this complementarity effectively: DF + CC Solver lifts Hard accuracy to 58.0\%, surpassing all prior reported baselines on this tier.
The best hybrid, DF + CC + QO, achieves 86.7\% overall accuracy — almost +9pp above the best SLR-Bench leaderboard result (o3: 77.8\%) — at 138.8M tokens (\$16.19), compared to \$207.24 for o3.
In contrast to PBEBench, all systems predict \emph{simpler} rules than ground truth ($\Delta < 0$), reflecting the abundance of shorter equivalent Prolog forms; symbolic solvers undershoot most aggressively, consistent with their ascending-complexity search strategy.

\textbf{Solver induction ablations.}
We study how the trace dataset composition affects induced solver quality (\S\ref{app:solver_ablations}). 
Removing reasoning traces (no-CoT) significantly degrades performance, especially on harder instances (e.g. PBEBench-Hard accuracy drops from 74.7\% to 24.8\%), indicating that final programs alone are insufficient supervision for learning robust solvers. 
In contrast, reducing the number of CoT examples has a smaller and less consistent effect: a 48-example CoT solver achieves performance comparable to some 100-example runs, while 12 examples is clearly insufficient. 
Notably, performance differences across solvers induced on the same trace dataset (100-examples + CoT) can be large (e.g., 53.4\%–79.2\% on Lite, 51.8\%–74.7\% on Hard), suggesting that solver quality is highly sensitive to the algorithmic strategy discovered during induction rather than trace dataset size alone.

\textbf{Real-world case study: forward reconstruction.}
We evaluate induced symbolic solvers on a real-world forward reconstruction task from historical linguistics, consisting of 3,077 proto-daughter language pairs with large variation in number of examples (ranging from 1 to 857) and an unseen IPA alphabet. 
Without any retraining, both CC and Qwen solvers achieve strong zero-shot performance ($\approx$ 70\% accuracy), demonstrating robustness to distribution shift beyond synthetic PBEBench tasks. 
Increasing the maximum cascade length from 20 to 100 yields only modest gains (e.g., 69.0\%→70.2\% for CC), suggesting that performance is not primarily limited by search depth.  
However, one independently induced solver (the unique-op permutation strategy in \S\ref{app:solver_ablations}) achieves only 4\% accuracy under distribution shift, illustrating that algorithmic brittleness, not data shift alone, limits generalization.
Combining solvers via union substantially improves robustness, reaching up to 80.1\% accuracy and demonstrating strong complementarity. 
Additional analysis, including compression experiments and qualitative examples, is provided in Appendix~\ref{app:real_fr}.
\section{Discussion}
\textbf{Efficiency under saturated performance.}
On PBEBench-Lite, accuracy is already near saturation for LLM-based methods, leaving limited room for improvement. 
In this regime, the primary benefit of solver induction is improved efficiency rather than accuracy. 
Hybrid inference substantially reduces LLM token usage, indicating that many tasks can be resolved through reusable symbolic structure instead of repeated LLM search. 
Among hybrids, the QO-based approach achieves Pareto efficiency, matching LLM accuracy while reducing both token usage and total cost.
The CC-based hybrid yields a marginal additional accuracy gain at comparable cost, as one-time solver construction overhead (\$2.00 for CC, \$0.85 for QO) is negligible relative to per-task inference savings.
The token reductions are proportional to solver coverage and independent of per-token pricing, suggesting the efficiency gains scale to any deployment context.
Beyond tokens, solver inference completes in under 5 minutes for all 1000+ tasks (pure Python, no LLM calls), compared to $\sim$2 days for DF or BoK at 8 parallel workers — a wall-clock gap of over 500$\times$ (Appendix~\ref{sec:appendix:computational_environment}).
Overall, these results indicate that even lightweight solvers suffice in saturated regimes, while more powerful solvers become critical on harder tasks.

\textbf{Symbolic solvers as primary baselines on Hard tasks.}
On PBEBench-Hard, the performance landscape inverts: LLM test-time scaling becomes the weaker baseline while symbolic solvers emerge as the primary competitive systems.
BoK achieves only 68.4\% at substantial cost (\$57.83), whereas induced symbolic solvers exceed this at zero per-task LLM cost (QO: 74.7\%, CC: 69.7\%), and the All Symbolic union reaches 84.7\%.
This inversion reflects the fundamental mismatch between independent LLM sampling and long-horizon cascade tasks: at cascade lengths 14--17, BoK collapses to below 47\% while symbolic solvers maintain 55--70\% accuracy through structured search.
Hybrid inference amplifies this advantage, with BoK + All Symbolic reaching 85.8\% while reducing token cost by 78\%.
The large complexity overshoot ($\Delta \approx +5$--$+8$) on Hard relative to Lite ($\Delta \approx +2$) indicates that inductive search at long cascade lengths favours valid-but-verbose solutions; ensembling partially corrects this by selecting simpler candidates across diverse solver outputs.
Notably, hybrid mean reward on Hard (e.g., BoK+CC: 0.9508) is lower than the standalone CC Solver (0.9873): when the solver fails, the LLM fallback is invoked independently and its partial-credit answers on hard tasks score lower than the solver's near-misses.
The same pattern holds on SLR-Bench, where the Hard tier exposes a sharp divide between LLM and symbolic approaches.
BoK collapses to 17.2\% on Hard while the CC Solver reaches 46.8\% — matching frontier models (o3, GPT-5) at zero inference cost.
Notably, the direct coding-agent baseline, despite consuming the most tokens of any system, achieves the lowest hard-tier performance, illustrating that per-task agentic effort does not substitute for the structured inductive search that symbolic solvers perform.
Hybrid methods combine both strengths: DF handles the Basic and Easy tiers near-perfectly while the symbolic solver provides hard-tier robustness that LLM search alone cannot reach, yielding the best overall and hard-tier results across all benchmarks evaluated.

\textbf{Solver induction ablations} confirm that reasoning traces are critical — removing CoT causes large accuracy drops especially on Hard tasks — and reveal that variance across induction runs (yielding qualitatively distinct algorithms) dominates the effect of dataset size, suggesting solver induction is a search problem over algorithmic space rather than a data-scaling problem.

\textbf{Generalization to real forward reconstruction.}
Most induced solvers transfer zero-shot to real IPA data (Appendix~\ref{app:real_fr}), but compression analysis reveals a gap between unconstrained accuracy ($\approx$70--80\%) and compact-rule performance ($\approx$30\%), suggesting current reward functions favor per-example patching over discovering generalizable structure.
Although the solvers do recover some plausible sound laws, this tendency is a formidable limitation for linguistic applications.
\section{Related Work}

\textbf{Neurosymbolic Reasoning with LLMs.}
A broad line of work improves LLM reasoning by coupling neural generation with symbolic execution, external tools, or explicit verification. PAL \citep{gao2023pal} delegates intermediate computations to executable programs, which can reduce arithmetic and logical errors. Follow-up work by \citet{kabra2024program} shows that program-aided reasoners can also be better calibrated than text-only chain-of-thought prompting. ReAct \citep{yao2022react} extends this paradigm to interleaved reasoning and acting during tool use. ReWOO \citep{xu2023rewoo} further separates reasoning from observation-heavy execution to reduce tool-calling overhead. CodeAct \citep{wang2024executable} treats executable Python code as a unified action space for LLM agents, allowing actions to be executed and revised through an interpreter over multiple turns. More recent analyses such as TIR \citep{zhao2025dissecting} study when tool-integrated reasoning helps in practice and when the added orchestration primarily increases latency or compute, and cost-aware evaluations make this trade-off even more explicit \citep{ghoshal2026tools}. LLM-PV \citep{singhal2025llm} uses an LLM as a proposal prior over programs and validates candidate programs by execution on held-out examples. Our Best-of-$K$ (BoK) baseline is closest in setup to this proposal-and-selection view: both sample candidate programs from an LLM and use execution-based signals to select among them. However, this line of work largely treats execution as an online verifier, tool-use substrate, or feedback signal for individual LLM-generated trajectories. In contrast, \methodname{} uses execution traces offline to induce a reusable symbolic solver, which is then deployed at test time before falling back to LLM search only on residual cases.

\textbf{Program Induction with LLMs.}
Work on program induction from examples predates LLMs. DeepCoder \citep{balog2016deepcoder} learns to guide search over a DSL from input-output examples, while RobustFill \citep{devlin2017robustfill} studies neural program induction for string transformations under noisy I/O. BUSTLE \citep{odena2020bustle} further shows that learned models can guide bottom-up search over programs from input-output examples by prioritizing promising compositions of intermediate values during synthesis. In the LLM era, \citet{li2024programming} show that off-the-shelf models remain weak PBE solvers but can become competitive after fine-tuning, especially in-distribution. HYSYNTH \citep{barke2024hysynth} proposes a hybrid approach in which LLM completions are distilled into a task-specific, context-free surrogate model that then guides symbolic program synthesis. 
\citet{verbruggen2025execution} propose execution-guided within-prompt search for PBE, where the model iteratively samples, combines, and annotates candidate code with execution results to carry out search within a single prompt.
In historical linguistics \citet{naik2025programming} cast sound law induction as PBE and explore synthetic training data for this setting. PBEBench \citep{naik2025pbebench} and SLR-Bench \citep{helff2025slr} provide executable, procedurally generated reasoning benchmarks that make compositional induction and scaling behavior more measurable. In a complementary application setting, PBE Meets LLM \citep{zhang2025pbe} studies tabular transformations with a hybrid solver-first pipeline, while Case2Code \citep{shao2025case2code} scales induction from program behavior into a synthetic data source for code models. What remains less explored is whether the search behavior of an LLM itself can be compiled into a standalone solver that amortizes inference across tasks.
\methodname{} addresses this gap by using a coding agent to induce symbolic solvers from LLM reasoning traces, then combining the induced solver with BoK or DF search in a hybrid test-time pipeline.
We also discuss additional work related to library learning in Appendix~\ref{app:more_related_work}.

\vspace{-5pt}
\section{Conclusion and Future Work}
\label{sec:conclusion}
\vspace{-5pt}
We show that LLM reasoning traces can be compiled into reusable symbolic solvers that are strong standalone systems, match or exceed frontier-model performance on hard tasks at zero per-task inference cost, and substantially reduce token usage as hybrids — Pareto-dominating direct per-task agentic effort.
The induced solvers are complementary to LLM search: they excel precisely where LLMs fail (long-horizon cascades, hard Prolog tiers) while LLMs cover the remaining tasks, and the best hybrids set new state-of-the-art results on both PBEBench and SLR-Bench.
Most solvers also generalize to an unseen IPA alphabet and real linguistic data without any source code/algorithmic changes, though compact-rule performance ($\approx$30\%) lags behind unconstrained accuracy ($\approx$70\%), suggesting solvers favor verifier-satisfying solutions over generalizable structure.
Future work could explore more extensive search over solver induction runs to optimize for robustness to domain shift, computational efficiency, or program complexity (the latter being directly relevant for linguistic interpretability); iterative solver improvement via targeted coding-agent debugging sessions; and automatic selection or mixture of induced solvers based on task features.



\bibliographystyle{plainnat}
\bibliography{custom}

\appendix
\section{Broader Impact}
\label{sec:appendix:broader_impact}

This work studies the induction of symbolic solvers from LLM reasoning traces and their combination with LLM test-time search for program synthesis tasks.
The contributions are primarily methodological: we introduce a framework for coding-agent-induced solver induction and demonstrate its application on two benchmarks in program-by-example synthesis and inductive logic programming.

\paragraph{Positive impacts.}
Reducing the per-task LLM cost of program synthesis---through reusable symbolic solvers that replace repeated LLM inference---lowers the computational and financial barrier to deploying capable synthesis systems, potentially improving accessibility.
Improved program synthesis tools have broad beneficial applications in data transformation, programming assistance, and scientific discovery (e.g., in formal linguistics, as demonstrated in our case study).

\paragraph{Potential negative impacts.}
We do not foresee direct negative societal impacts from this work.
The string-replacement and Prolog-rule DSLs studied here are highly constrained and domain-specific, with no straightforward path to harmful applications such as disinformation generation, surveillance, or privacy violations.
More generally, improvements in program synthesis efficiency could in principle accelerate the development of automation tools with broader downstream effects; however, this path is indirect and not specific to our method.
The coding agents used in solver induction (LLMs instructed to write code) could in principle produce solvers that behave incorrectly or unpredictably, but our evaluation framework includes a verifier that explicitly checks correctness on held-out examples, limiting the risk of undetected failure.

\paragraph{Fairness and privacy.}
The benchmarks used in this work are synthetic or drawn from published linguistics datasets and contain no personal data.
No demographic information is collected or modelled; no deployment decisions affecting individuals are made.
We therefore identify no fairness or privacy concerns specific to this work.

\begin{figure*}[t]
\centering
\includegraphics[width=0.9\textwidth]{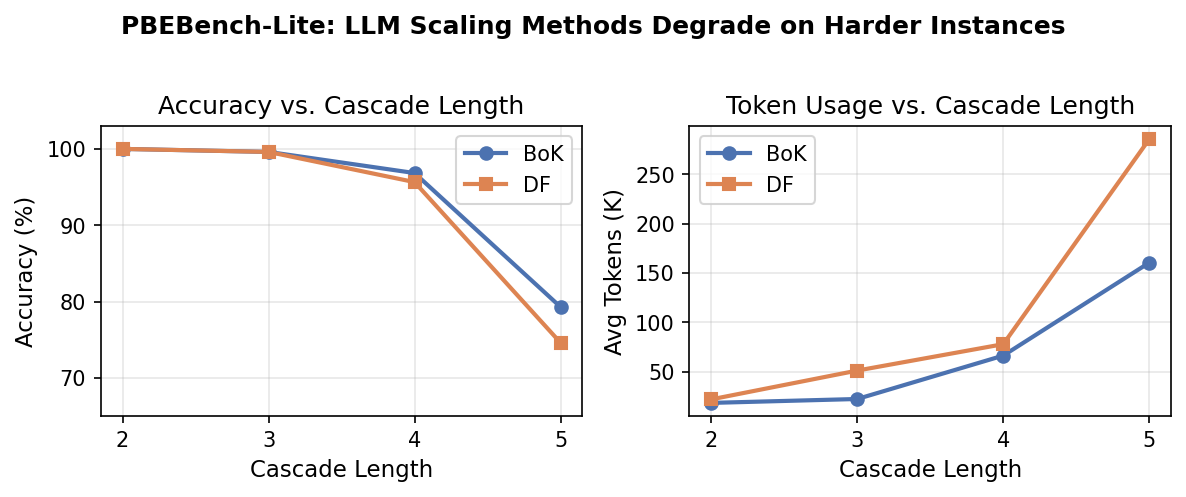}
\caption{\textbf{LLM scaling methods degrade on harder PBEBench-Lite instances.}
Accuracy (left) and average token usage (right) of Best-of-K (BoK) and Direct Feedback (DF) scaling strategies applied to gpt-oss-120b, broken down by cascade length on PBEBench-Lite.
Both methods solve cascade-length-2 tasks perfectly but collapse at length 5: BoK accuracy drops to 79.3\% and DF to 74.6\%, while token usage explodes --- DF consumes an average of 285K tokens per task at length 5 ($13{\times}$ more than at length 2), and BoK consumes 160K tokens ($9{\times}$ more).
This shows that harder instances trigger more LLM attempts, compounding both cost and error.}
\label{fig:pbebench_lite_by_cl}
\end{figure*}
\section{Limitations}
\label{sec:limitations}

\paragraph{Scope of evaluation.}
Our experiments are conducted on two program synthesis domains: string-replacement cascades (PBEBench) and binary train-classification rules (SLR-Bench).
Both domains use highly structured, verifiable DSLs.
It is not clear whether solver induction generalizes to domains with more complex DSLs, open-ended output spaces, or without a fast and exact verifier.
The forward reconstruction case study (Appendix~\ref{app:real_fr}) provides initial evidence of zero-shot transfer, but that domain shares structural properties with PBEBench.

\paragraph{Run-to-run variance in solver induction.}
Solver quality varies substantially across independent induction runs on identical demonstrations (PBEBench-Lite accuracy 53--79\% for three identical 100-example CoT runs with Qwen3.6-35B-A3B).
This variance reflects qualitatively different algorithms discovered by the coding agent across trajectories and is not reducible by fixing a random seed, as the agent's exploration is governed by its own sampling.
Reported results for the QO Solver use the best single run (selected by performance on the induction demo set, not the test set); ensemble results over multiple runs are provided but depend on the number of induction attempts.
The CC Solver was induced in a single session per benchmark; no multi-run variance estimate is available for it.
This variance could confound the CoT and dataset-size ablations in Appendix~\ref{app:solver_ablations}, since single ablation runs are compared against multiple CoT runs.
However, we note that the performance reductions under the most extreme ablations — removing CoT entirely (Hard accuracy 24.8\% vs.\ 51.8--74.7\% for CoT runs) — exceeds the observed run-to-run algorithmic variance (up to 22.9pp on Hard), providing evidence that these ablation effects reflect some genuine differences rather than sampling noise alone.

\paragraph{Dependence on coding agent capability.}
Solver quality is sensitive to the capability of the coding agent used for induction: our ablations show that removing chain-of-thought supervision dramatically degrades performance (e.g., Hard accuracy drops from 74.7\% to 24.8\% for the no-CoT QO Solver), and that 12 demonstration examples are insufficient for reliable induction.
Weaker coding agents or smaller trace budgets may not successfully induce useful solvers, making the approach contingent on access to sufficiently capable code-generation models.

\paragraph{Complexity bias.}
Induced solvers can produce programs of higher complexity than ground truth (positive $\Delta$ throughout on PBEBench), particularly on PBEBench-Hard.
While accuracy is unaffected, over-complex programs are less interpretable and may generalize poorly to unseen distributions.
The compression analysis in the forward reconstruction case study (Appendix~\ref{app:real_fr}) reveals a large gap between unconstrained accuracy ($\approx$70--80\%) and compact-rule performance ($\approx$30\%), suggesting that the verifier-based reward function favours solutions that fit the given examples rather than those that capture generalizable structure.

\paragraph{Token savings depend on solver coverage.}
The efficiency gains from hybrid inference are proportional to the fraction of tasks the symbolic solver solves perfectly.
Solvers with lower coverage provide smaller token reductions.
On PBEBench-Lite, where the CC Solver covers 80.4\% of tasks perfectly, token savings are substantial ($\sim$33\% for BoK + CC); on harder tasks or weaker solvers, savings are smaller.
Practitioners should therefore evaluate solver coverage on their target distribution before expecting comparable efficiency gains.

\paragraph{Single-model LLM baselines.}
All BoK and DF baselines use a single model (gpt-oss-120b).
Results may not generalize to other reasoning models, and the relative advantage of symbolic solvers may vary with model capability.
In particular, as the frontier model's performance on these benchmarks approaches saturation, the accuracy benefit of symbolic solvers diminishes, and efficiency becomes the primary motivation.
\section{Computational Environment}
\label{sec:appendix:computational_environment}

This section documents the hardware, software, and per-experiment compute budgets required to reproduce all results in the paper.

\subsection{Hardware}

Table~\ref{tab:hardware} summarises the hardware used for each experimental role.

\begin{table}[!tbh]
\centering
\caption{Hardware configuration by experimental role.}
\label{tab:hardware}
\resizebox{\textwidth}{!}{
\begin{tabular}{lll}
\toprule
\textbf{Role} & \textbf{Hardware} & \textbf{Notes} \\
\midrule
Open-source LLM inference (vLLM) & 2 $\times$ NVIDIA A100 80\,GB & Internal cluster; serves \texttt{gpt-oss-120b} and \texttt{Qwen3.6-35B-A3B} \\
Claude Code solver induction & Anthropic API (\texttt{claude-sonnet-4-6}) & Interactive CLI session; no local GPU required \\
SLR-Bench verifier & CPU & SWI-Prolog subprocess ($\sim$300\,ms per rule) \\
Eval / analysis scripts & CPU & Python (stdlib, Pandas, Matplotlib) \\
\bottomrule
\end{tabular}
}
\end{table}

All self-hosted inference was executed on an internal university cluster. \textbf{Reported dollar costs are reference estimates computed using public API pricing} (DeepInfra for \texttt{gpt-oss-120b}; AtlasCloud for \texttt{Qwen3.6-35B-A3B}) \textbf{for comparison purposes only}; actual experiments were run on the cluster at no direct monetary cost.

\subsection{Software}

Table~\ref{tab:software} lists the software versions used across all experiments.

\begin{table}[!tbh]
\centering
\caption{Software versions.}
\label{tab:software}
\begin{tabular}{ll}
\toprule
\textbf{Component} & \textbf{Version} \\
\midrule
Python & 3.11 \\
vLLM & 0.8.x \\
openhands-sdk & 1.16.1 \\
Apptainer image & \texttt{python:3.11-slim} + numpy / scipy / sympy / evaluate / datasets \\
SWI-Prolog & 9.x \\
Anthropic Python SDK & Latest (\texttt{claude-sonnet-4-6}) \\
\bottomrule
\end{tabular}
\end{table}

\subsection{Per-Experiment Compute}

Reference pricing used for cost estimates:
\begin{itemize}
    \item \textbf{gpt-oss-120b}: \$0.039\,/\,M input tokens, \$0.190\,/\,M output tokens (reasoning billed as output) --- DeepInfra public pricing
    \item \textbf{Qwen3.6-35B-A3B}: \$0.1612\,/\,M input tokens, \$0.9653\,/\,M output tokens --- AtlasCloud public pricing
\end{itemize}

Token counts are exact, taken from provider usage fields logged in every output JSONL\@. ``M tok'' denotes millions of tokens (input + output combined).

\paragraph{PBEBench-Lite (1008 tasks, \texttt{max\_programs}~=~5)}
Token and cost breakdown is shown in Table~\ref{tab:compute_lite}.

\begin{table*}[!tbh]
\centering
\caption{Per-experiment token and reference cost breakdown for PBEBench-Lite.}
\label{tab:compute_lite}
\resizebox{\textwidth}{!}{
\begin{tabular}{llrrrl}
\toprule
\textbf{Experiment} & \textbf{Model} & \textbf{Tasks} & \textbf{M tok} & \textbf{Ref.\ cost} & \textbf{Notes} \\
\midrule
DF-32 & gpt-oss-120b & 1008 & 111.1 & \$16.74 & 32 attempts max, single-turn per attempt \\
BoK-32 & gpt-oss-120b & 1008 & 68.0 & \$12.20 & 32 independent samples \\
QO Agent / DirectSolve & Qwen3.6-35B-A3B & 1008 & 395.3 & \$85.38 & Complete; avg 11.1 steps/task (max 100) \\
CC Solver (build) & claude-sonnet-4-6 (CC CLI) & 1 session & --- & \$2.00 & One-time; $\sim$30\,min, exact cost from tracked run \\
QO Solver run 1 (build) & Qwen3.6-35B-A3B & 1 session & 4.46 & \$0.79 & 76 turns, 201\,min \\
QO Solver run 2 (build) & Qwen3.6-35B-A3B & 1 session & 4.82 & \$0.85 & 72 turns, 34\,min \\
QO Solver run 3 (build) & Qwen3.6-35B-A3B & 1 session & 6.24 & \$1.10 & 80 turns, 44\,min \\
QO Solver (100ex, no CoT, build) & Qwen3.6-35B-A3B & 1 session & 7.81 & \$1.34 & 102 turns, 44\,min (ablation) \\
QO Solver (48ex + CoT, build) & Qwen3.6-35B-A3B & 1 session & 4.18 & \$0.74 & 82 turns, 26\,min (ablation) \\
QO Solver (12ex + CoT, build) & Qwen3.6-35B-A3B & 1 session & 1.65 & \$0.30 & 49 turns, 15\,min (ablation) \\
\bottomrule
\end{tabular}
}
\end{table*}

Solver inference incurs zero per-task LLM cost (pure Python). Ensemble outputs are computed deterministically from individual outputs; no additional LLM calls are required.

\paragraph{PBEBench-Hard (1216 tasks, \texttt{max\_programs}~=~20)}
Token and cost breakdown is shown in Table~\ref{tab:compute_hard}.

\begin{table*}[!tbh]
\centering
\caption{Per-experiment token and reference cost breakdown for PBEBench-Hard.}
\label{tab:compute_hard}
\resizebox{\textwidth}{!}{
\begin{tabular}{llrrrl}
\toprule
\textbf{Experiment} & \textbf{Model} & \textbf{Tasks} & \textbf{M tok} & \textbf{Ref.\ cost} & \textbf{Notes} \\
\midrule
BoK-32 & gpt-oss-120b & 1216 & 332.1 & \$57.83 & Complete \\
QO Agent / DirectSolve & Qwen3.6-35B-A3B & 1216 & 1046.8 & \$224 & Complete; avg 861K tok/task; avg 18.9 steps/task \\
CC Solver (inference only) & --- & 1216 & 0 & \$2.00 build & Same solver as Lite; build $\sim$30\,min \\
QO Solver run 2 (inference only) & --- & 1216 & 0 & \$0.85 build & Same solver as Lite run 2 \\
\bottomrule
\end{tabular}
}
\end{table*}

\paragraph{SLR-Bench (1000 tasks)}
Token and cost breakdown is shown in Table~\ref{tab:compute_slr}.

\begin{table*}[!tbh]
\centering
\caption{Per-experiment token and reference cost breakdown for SLR-Bench.}
\label{tab:compute_slr}
\resizebox{\textwidth}{!}{
\begin{tabular}{llrrrl}
\toprule
\textbf{Experiment} & \textbf{Model} & \textbf{Tasks} & \textbf{M tok} & \textbf{Ref.\ cost} & \textbf{Notes} \\
\midrule
DF-32 & gpt-oss-120b & 1000 & 224.2 & \$17.43 & Complete \\
BoK-32 & gpt-oss-120b & 1000 & 225.3 & \$17.88 & Complete \\
QO Agent / DirectSolve & Qwen3.6-35B-A3B & 1000 & 1232.6 & \$243.41 & Complete; avg 1,233K tok/task; avg 14.8 steps/task \\
CC Solver (build) & claude-sonnet-4-6 (CC CLI) & 1 session & --- & \$4.01 & One-time; $\sim$30\,min, exact cost from tracked run \\
QO Solver run 1 (build) & Qwen3.6-35B-A3B & 1 session & 2.93 & \$0.51 & 68 turns, 25\,min \\
QO Solver run 2 (build) & Qwen3.6-35B-A3B & 1 session & 7.49 & \$1.28 & 84 turns, 47\,min \\
\bottomrule
\end{tabular}
}
\end{table*}

\subsection{Total Compute}

Table~\ref{tab:compute_total} summarises total reference costs across all reported experiments.

\begin{table}[!tbh]
\centering
\caption{Total reference compute costs across all reported experiments.}
\label{tab:compute_total}
\resizebox{\textwidth}{!}{%
\begin{tabular}{lr}
\toprule
\textbf{Category} & \textbf{Ref.\ cost} \\
\midrule
gpt-oss-120b inference (Lite DF + BoK, Hard BoK, SLR DF + BoK) & $\sim$\$122 \\
Qwen3.6-35B-A3B --- QO Agent / DirectSolve (Lite + Hard + SLR) & $\sim$\$552 \\
Qwen3.6-35B-A3B --- QO Solver induction (6 PBE + 2 SLR runs) & $\sim$\$7 \\
claude-sonnet-4-6 --- CC Solver induction (1 PBE + 1 SLR session) & $\sim$\$6 (\$2.00 PBE + \$4.01 SLR) \\
\midrule
\textbf{Total (reported experiments)} & \textbf{$\sim$\$687} \\
\bottomrule
\end{tabular}
}
\end{table}

All costs are reference figures based on public API pricing. Actual compute was executed on an internal cluster (2$\times$A100 80\,GB) at no direct monetary cost.

\subsection{Execution Time}

Table~\ref{tab:wallclock} reports approximate wall-clock times for each experimental phase.

\begin{table}[!tbh]
\centering
\caption{Approximate wall-clock times by experimental phase.}
\label{tab:wallclock}
\begin{tabular}{ll}
\toprule
\textbf{Experiment} & \textbf{Wall-clock time} \\
\midrule
QO Solver build (one run) & 15--201\,min (median $\sim$40\,min) \\
CC Solver build (one session) & $\sim$30\,min (PBE and SLR) \\
DF / BoK inference ($\sim$1000 tasks, 8 workers) & $\sim$2 days per benchmark \\
QO Agent / DirectSolve (1000 tasks, 8 workers) & 24--72 hours (hard-tier tasks $\sim$3.5M tok/task) \\
Symbolic solver inference (all tasks) & $<$5\,min (pure Python, no LLM calls) \\
SLR-Bench verifier & $\sim$300\,ms per rule; $\sim$5\,min for 1000 tasks \\
\bottomrule
\end{tabular}
\end{table}

\subsection{Parallelism}

DF, BoK, and QO Agent baselines are executed with 8 parallel workers via Python \texttt{ThreadPoolExecutor} (one task per thread). Solver induction (QO and CC) runs as a single sequential agent session per induction run; multiple runs were launched independently in parallel. Ensemble aggregation is single-threaded and deterministic, requiring no additional LLM calls.
\section{Method Details}

\subsection{Solver Building Prompts}
\label{app:prompts}

\begin{tcolorbox}[title=PBEBench Solver Prompt]
You are an expert in symbolic program induction.
\\
\\
Write a single Python file implementing a solver for a given Programming by Example (PBE) task.
\\
\\
\#\# Task
\\
\\
Write a Python-based symbolic program synthesizer that infers a transformation program from a set of ( input\_string, output\_string ) pairs.
\\
\\
You will be shown examples of an LLM solving similar tasks in @DEMOS\_PBEBENCH.json, including reasoning traces from both successful and unsuccessful attempts across easy and hard cases. Use these to understand the task structure and take inspiration from the problem-solving strategies, especially in cases where the LLM struggles.
\\
\\
Output: 
\\
- a Python solver file @SOLVER.py
\\
- a markdown file @SOLVER\_ALGORITHM.md explaining the algorithm

The solver should use the verifier defined in @rewards/pbebench.py to evaluate candidate programs. If no correct program is found, it should return the top-K highest scoring programs, where K is a parameter taken by the solver.

\#\# Requirements
\\
\\
* Output exactly one Python file and one markdown file. 
\\
* Use only the Python standard library.
\\
* No external data, APIs, or dataset-specific assumptions.
\\
* The solver must generalize across tasks in @DEMOS\_PBEBENCH.json

\#\# Interface

Implement:

def solve\_pbe(examples): 
\\
    """
\\
    \quad examples: list of (input\_string, output\_string)
\\
    returns: dict with at least:
\\
        - "success": bool
\\
        - "program": structured representation of the inferred transformation
\\
    """

\#\# Behavior

The solver should:
\\
* infer a program consistent with the examples and compatible with the verifier
\\
* use (not reimplement) the verifier to score candidate programs
\\
* prefer simple, compositional rules with low description complexity
\\
* follow the domain-specific language (DSL) defined in @DEMOS\_PBEBENCH.json
\\
* search over candidate transformations and select ones that match all examples
\\
* if no fully consistent program is found, return the top-K highest scoring programs
\\
* return structured programs or hypotheses that could be useful for downstream refinement if partially incorrect
\end{tcolorbox}

\begin{tcolorbox}[title=SLR-Bench Solver Prompt (1/2)]
\small
You are an expert in symbolic program induction.
\\
\\
Write a single Python file implementing a solver for a given SLR-Bench task.
\\
\\
\#\# Task
\\
\\
Write a Python-based symbolic program synthesizer that infers a Prolog rule from a set of (background\_facts, direction
\_label) pairs, where direction\_label is either "eastbound" or "westbound".
\\
\\
You will be shown examples of an LLM solving similar tasks in @DEMOS\_SLRBENCH.json, including reasoning traces from both successful and unsuccessful attempts across easy and hard cases. Use these to understand the task structure and take inspiration from the problem-solving strategies, especially in cases where the LLM struggles.
\\
\\
Output:
\\
- a Python solver file @SOLVER\_SLR.py
\\
- a markdown file @SOLVER\_SLR\_ALGORITHM.md explaining the algorithm
\\
\\
The solver should use the verifier defined in @rewards/slr\_bench.py to evaluate candidate rules. If no correct rule is found, it should return the top-K highest scoring rules, where K is a parameter taken by the solver.
\\
\\
\#\# Requirements
\\
\\
* Output exactly one Python file and one markdown file.
\\
* Use only the Python standard library.
\\
* No external data, APIs, or dataset-specific assumptions.
\\
* The solver must generalize across tasks in @DEMOS\_SLRBENCH.json
\\
\\
\#\# Interface
\\
\\
Implement:
\\
\\
def solve\_slr(examples):
\\
    """
\\
    examples: list of (facts\_string, direction\_label)
\\
              facts\_string  — space-separated Prolog ground facts for one train
\\
              direction\_label — "eastbound" or "westbound"
\\
    returns: dict with at least:
\\
        - "success": bool
\\
        - "program": Prolog rule string of the form "eastbound(T) :- Body."
\\
    """
\end{tcolorbox}

\begin{tcolorbox}[title=SLR-Bench Solver Prompt (2/2)]
\small
\#\# Domain-Specific Language
\\
\\
The rule must be a Prolog clause of the form `eastbound(T) :- Body.` where Body is a conjunction of
literals drawn from these predicates:
\\
\\
- `has\_car(Train, Car)` — Car is part of Train
\\
- `car\_num(Car, CarNumber)` — position of Car (positive integer)
\\
- `car\_color(Car, Color)` — Color $\in$ {red, blue, green, yellow, white}
\\
- `car\_len(Car, Length)` — Length $\in$ {short, long}
\\
- `has\_wall(Car, WallType)` — WallType $\in$ {full, railing}
\\
\\
Prefer rules with the fewest body literals (use `rule\_complexity()` from `rewards/slr\_bench.py`
to measure this).
\\
\\
\#\# Performance
\\
\\
Each call to the verifier (`judge.compute`) invokes SWI-Prolog as a subprocess and costs ~300ms.
Hard tasks can have hundreds of thousands of candidate rules. Minimize the number of verifier
calls — use the examples to prune the candidate space in Python before invoking the verifier.
\\
\\
\#\# Behavior
\\
\\
The solver should:
\\
* infer a rule consistent with the examples and compatible with the verifier
\\
* use (not reimplement) the verifier to score candidate rules
\\
* prefer simple rules with the fewest body literals
\\
* follow the domain-specific language defined above and in @DEMOS\_SLRBENCH.json
\\
* search over candidate rules and select ones that correctly classify all examples
\\
* if no fully consistent rule is found, return the top-K highest scoring rules
\\
* return structured rules or hypotheses that could be useful for downstream refinement if partially incorrect
\end{tcolorbox}
\section{More Experimental Details}
\label{sec:appendix:experimental_details}

\subsection{Token accounting.}
\label{sec:appendix:experimental_details:token_accounting}
Token costs for \texttt{gpt-oss-120b} are computed using open-router pricing (\$0.039/M input, \$0.19/M output; reasoning tokens billed as output). 
Solver construction costs for QO are computed from OpenHands trajectories using native tokenization for \texttt{Qwen3.6-35B-A3B}. 
For \texttt{Qwen3.6-35B-A3B} as well, we use open-router pricing (\$0.1612/M input, \$0.9653/M output; reasoning tokens billed as output). 
Symbolic solvers incur zero per-task inference cost; construction costs are small ($\sim\$0.3$--\$1.3 per solver) and amortize after a handful of tasks. 
For hybrid methods, reported token usage includes only fallback LLM calls, reflecting effective cost under reuse.
\section{More Results}
\label{sec:appendix:more_results}

\subsection{Additional Details: PBEBench-Lite}
\label{app:lite}

\paragraph{Search space.}
Each PBEBench program is a \texttt{replace(A,B)} operation where the predicate $A$ has length 1--3 and the replacement $B$ has length 0--3. 
For an alphabet of size $V$, this gives
\[
(V + V^2 + V^3)(1 + V + V^2 + V^3)
\]
possible programs. 
A cascade of length $k$ therefore induces a search space exponential in $k$.

\begin{table}[t]
\centering
\caption{PBEBench search-space sizes. Each row reports the number of possible \texttt{replace(A,B)} programs and the resulting worst-case cascade search space.}
\smallskip
\label{tab:pbebench_search_space}
\begin{tabular}{lcccc}
\toprule
Setting & $V$ & Programs & Max cascade & Search space \\
\midrule
Lite (per-task avg.) & 13 & $5.66{\times}10^6$ & 5 & $\approx 5.8{\times}10^{33}$ \\
Lite (global vocab.) & 17 & $2.72{\times}10^7$ & 5 & $\approx 1.5{\times}10^{37}$ \\
Hard (global vocab.) & 52 & $2.06{\times}10^{10}$ & 20 & $\approx 1.8{\times}10^{206}$ \\
\bottomrule
\end{tabular}
\end{table}

Even the Lite search space is far beyond exhaustive enumeration: at $10^9$ program evaluations per second and $\approx 10^7$ seconds per year, the per-task-average Lite setting would require roughly $10^{17}$ years; the Hard setting is cosmologically larger ($\approx 10^{190}$ years).
The induced symbolic solvers avoid this enumeration by extracting candidate predicates directly from input-output differences, pruning replacements to a small candidate set, and applying greedy or beam search over cascades. 
Thus, they evaluate hundreds to thousands of candidates per task rather than enumerating the full combinatorial space.

\begin{figure*}[!tbh]
    \centering
    \includegraphics[width=\textwidth]{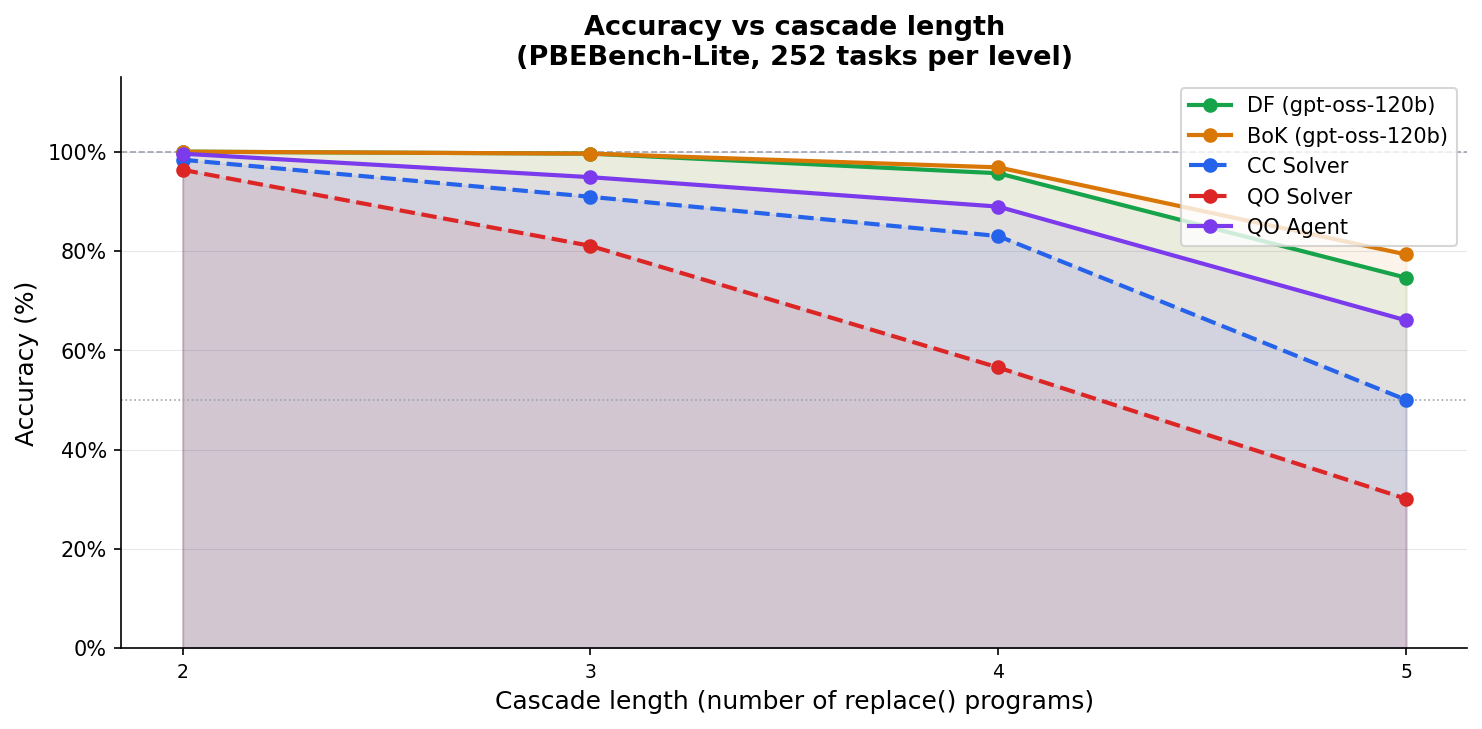}
    \caption{\textbf{PBEBench-Lite Accuracy vs Cascade Length:} LLM scaling strategies dominate, even on harder instances, even though the overall data is relatively simpler compared to PBEBench-Hard.}
    \label{fig:pbebench_lite_accuracy}
\end{figure*}

\begin{figure*}[!tbh]
    \centering
    \includegraphics[width=\textwidth]{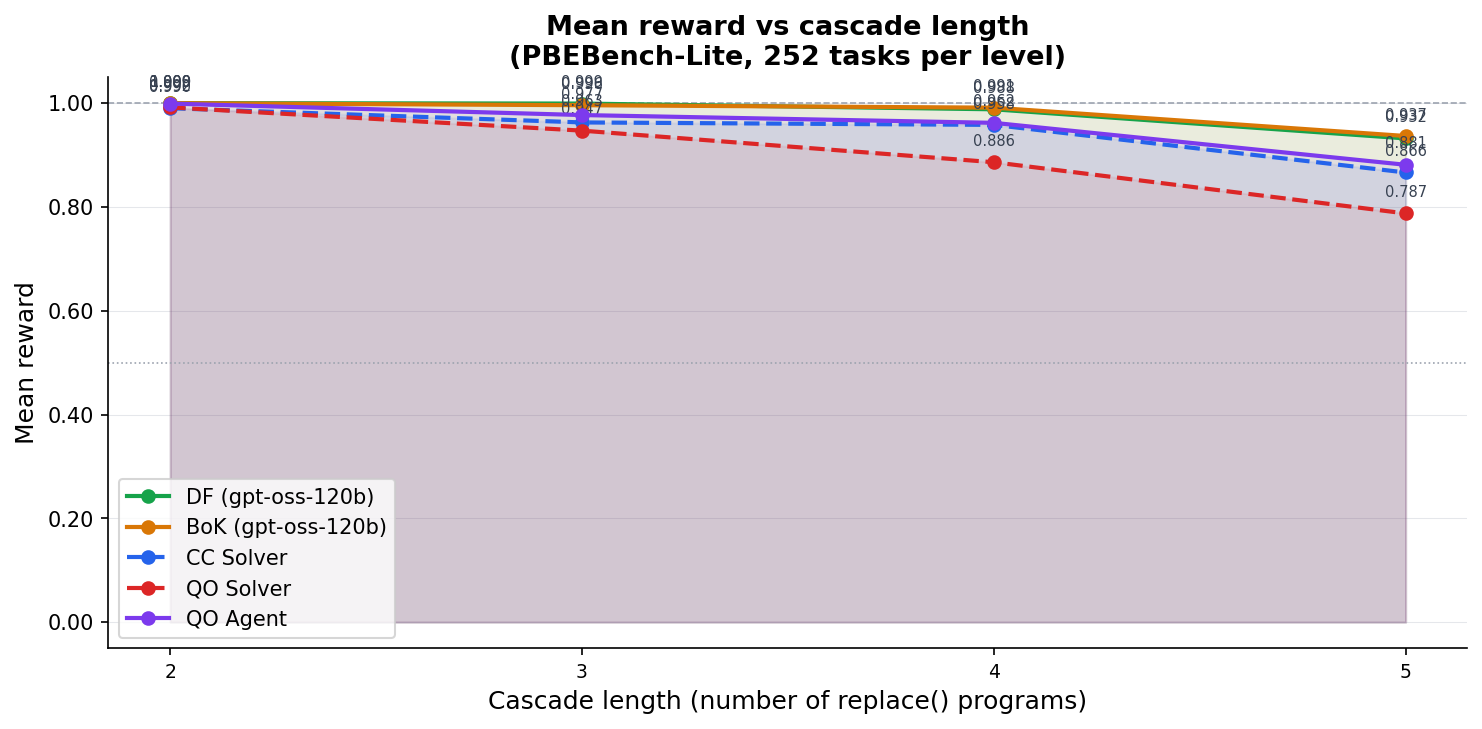}
    \caption{\textbf{PBEBench-Lite mean reward vs.\ cascade length.} The partial-credit gap between solvers and LLM scaling methods is smaller than the full-accuracy gap, confirming that solver failures are predominantly near-misses.}
    \label{fig:pbebench_lite_mean_reward}
\end{figure*}

\begin{figure*}[!tbh]
    \centering
    \includegraphics[width=\textwidth]{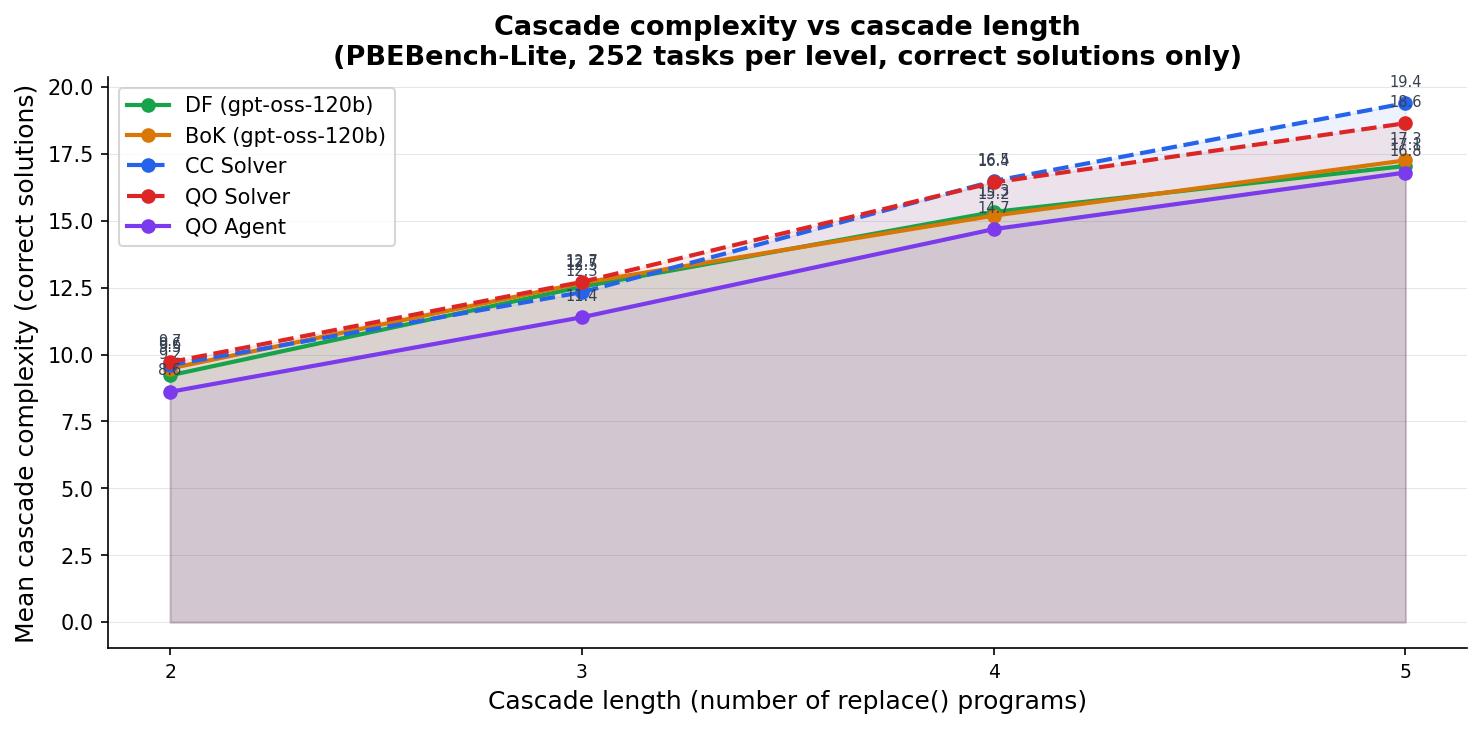}
    \caption{\textbf{PBEBench-Lite cascade complexity vs.\ cascade length.} Symbolic solvers overshoot ground-truth complexity ($\Delta \approx +3$); the direct coding agent (QO Agent) achieves the lowest complexity through multi-turn refinement.}
    \label{fig:pbebench_lite_complexity}
\end{figure*}

\paragraph{Breakdown by cascade length.} For PBEBench-Lite even for the symbolic solvers the performance degrades sharply at cascade length of 5, indicating difficulty in handling ordering interactions under tight program budgets.
We also show comparisons of the accuracy (full correctness), mean reward obtained (partial correctness) and complexity of the generated programs for the CC and QO solvers, compared to the scaling methods (DF and BoK) and the QO direct coding agent baseline in Figures~\ref{fig:pbebench_lite_accuracy}, \ref{fig:pbebench_lite_mean_reward} and \ref{fig:pbebench_lite_complexity}.
The figures show that due to the simpler nature of the PBEBench-Lite dataset it is nearly saturated by the scaling strategies and they tend to dominate even for harder cases.
However the symbolic solvers still lead to token efficiency and the performance gap is smaller when considering partial correctness via the mean reward.
Additionally directly solving the problems with a coding agent performs worse compared to simple LLM scaling but is better than the solvers.
For complexity the solvers tend to induce more complex programs than LLM scaling but the coding agent due to its multi-turn refinement strategy ends up achieving the lowest complexity/simplest programs.

\begin{table}[t]
\centering
\caption{CC and QO solver accuracy and mean reward by cascade length on PBEBench-Lite.}
\smallskip
\label{tab:lite_by_cl}
\begin{tabular}{rrccccc}
\toprule
\textbf{CL} & \textbf{N} & \textbf{Acc\% (CC)} & \textbf{Reward (CC)} & \textbf{Acc\% (QO)} & \textbf{Reward (QO)} \\
\midrule
2 & 246 & 98.4 & 0.990 & 96.3 & 0.992 \\
3 & 253 & 90.9 & 0.963 & 81.0 & 0.947 \\
4 & 253 & 83.0 & 0.958 & 56.5 & 0.886 \\
5 & 256 & 50.0 & 0.866 & 30.1 & 0.788 \\
\bottomrule
\end{tabular}
\end{table}

\subsection{Solver Construction Ablations and Analysis}
\label{app:solver_ablations}

\paragraph{Setup.}
We analyze how the composition of the demos file (number of examples and presence of reasoning traces) affects solver quality on PBEBench. 
We induce six solvers via the Qwen3.6-35B-A3B + OpenHands agent, varying either the demos file or nothing at all while keeping all other settings fixed (same prompt, verifier, and seed-42 balanced sampling across success/failure $\times$ easy/hard quadrants). Token costs are measured using the native Qwen tokenizer. 

\paragraph{Performance.}
Table~\ref{tab:solver_ablation_performance} denotes the performance of the induced solver for various trace dataset and solver inductions settings.

\begin{table}[h]
\centering
\small
\caption{Solver performance across trace dataset configurations. The Lite Acc and Hard Acc columns denote the accuracy on PBEBench-Lite and PBEBench-Hard while the Lite ES and Hard ES columns denote the edit similarity scores, with Algorithm summarizes the approach of the solver. Demo accuracy is evaluated on the PBEBench trace dataset instances the solver was fitted on.}
\label{tab:solver_ablation_performance}
\smallskip
\resizebox{\textwidth}{!}{
\begin{tabular}{lrrrrrrrl}
\toprule
Demos & Run & Lite Acc. & Lite Edit Sim. & Hard Acc. & Hard Edit Sim. & Demo Acc. & Demo Reward & Algorithm \\
\midrule
100 examples + CoT & 1 & 53.4 & 78.6 & 58.9 & 94.5 & 71.4 & 0.9875 & greedy + multi-pass residual fixing \\
100 examples + CoT & 2 & 65.7 & 87.4 & 74.7 & 96.8 & 87.9 & 0.9930 & safety-first greedy + 2-step lookahead \\
100 examples + CoT & 3 & \textbf{79.2} & \textbf{96.7} & 51.8 & 90.0 & 75.8 & 0.8774 & unique-op permutations + greedy + 2-op sequences \\
100 examples, no CoT & 1 & 42.1 & 67.6 & 24.8 & 82.9 & 48.4 & 0.9633 & beam search + heuristic scoring \\
48 examples + CoT & 1 & 55.7 & 78.6 & \textbf{76.2} & 96.5 & 93.5 & 0.9930 & multi-start greedy + permutation reorder \\
12 examples + CoT & 1 & 47.7 & 80.2 & 50.4 & 94.3 & 58.3 & 0.9917 & adaptive beam search \\
\bottomrule
\end{tabular}
}
\end{table}

\paragraph{Solver construction token cost.} The token costs for each solver estimated using the OpenHands trajectories with the native tokenizer and open-router pricing are shown in Table~\ref{tab:solver_construction_cost} for each PBEBench specific solver. The cost tends to be in a small range of \$0.3-1.34 showing the affordability of symbolic solver induction.

\begin{table}[h]
\centering
\small
\caption{Solver construction token costs at AtlasCloud pricing for Qwen3.6-35B-A3B. Each turn accumulates the full context as input; retry bloat is excluded.}
\label{tab:solver_construction_cost}
\smallskip
\begin{tabular}{lrrrrrr}
\toprule
Demos & Run & Turns & Input tokens & Output tokens & Total tokens & Cost (\$) \\
\midrule
100 examples + CoT & 1 & 76 & 4{,}379{,}412 & 82{,}982 & 4{,}462{,}394 & 0.79 \\
100 examples + CoT & 2 & 72 & 4{,}725{,}111 & 93{,}839 & 4{,}818{,}950 & 0.85 \\
100 examples + CoT & 3 & 80 & 6{,}122{,}331 & 114{,}529 & 6{,}236{,}860 & 1.10 \\
100 examples, no CoT & 1 & 102 & 7{,}721{,}064 & 93{,}617 & 7{,}814{,}681 & 1.34 \\
48 examples + CoT & 1 & 82 & 4{,}103{,}766 & 76{,}919 & 4{,}180{,}685 & 0.74 \\
12 examples + CoT & 1 & 49 & 1{,}607{,}936 & 43{,}796 & 1{,}651{,}732 & 0.30 \\
SLR, run 1 & 68 & -- & 2{,}888{,}522 & 42{,}666 & 2{,}931{,}188 & 0.51 \\
SLR, run 2 & 84 & -- & 7{,}404{,}083 & 90{,}826 & 7{,}494{,}909 & 1.28 \\
\bottomrule
\end{tabular}

\end{table}

\paragraph{Qualitative trajectory analysis.}
Each OpenHands run produces a different solver algorithm despite identical inputs, indicating exploration over algorithmic space.

\textbf{100 examples + CoT, run 1 (Greedy + residual fixing).}
Extracts edit regions and generates candidates via substitution, split edits, and context extension. Greedy selection maximizes fixes while minimizing regressions, followed by residual repair passes. Captures multi-step edits through split candidates but often produces overly complex cascades.

\textbf{100 examples + CoT, run 3 (Unique-op permutations).}
Separates forced vs optional operations and enumerates permutations over forced ones, followed by greedy and combinatorial search over optional candidates. Explicitly searches 2-step sequences for hard pairs. Highly effective on Lite but degrades on Hard due to fewer forced constraints.

\textbf{100 examples + CoT, run 2 (Safety-first greedy + lookahead).}
Applies a hard constraint to avoid modifying unchanged examples, combined with 2-step lookahead for interaction effects. Produces a tighter search space and strong overall performance.

\textbf{100 examples, no CoT (Beam search).}
Defaults to a generic beam search with heuristic scoring. Lacks structural insights (e.g., edit regions or ordering), leading to poor scaling on longer cascades.

\textbf{48 examples + CoT (Multi-start + reorder).}
Separates construction and ordering, using permutation search to explicitly handle interaction effects. Achieves the best Hard performance among Qwen runs.

\textbf{12 examples + CoT (Adaptive beam).}
Uses adaptive beam search with diversity constraints but lacks sufficient signal to discover structural insights, resulting in weaker performance.

\paragraph{Claude Code PBEBench solver.}
The Claude Code solver uses a two-phase safe/unrestricted beam search with candidate extraction from \texttt{difflib.SequenceMatcher}. Phase 1 enforces a safety constraint by restricting candidates to patterns that do not occur in already-correct inputs, while Phase 2 relaxes this constraint for rare cases where a correct replacement must touch such a pattern. Candidates are generated dynamically from intermediate states, so later beam depths can discover feeding and bleeding interactions induced by earlier programs. Context-extended patterns improve specificity and reduce unintended replacements.
It explicitly models intermediate states, enabling discovery of ordering interactions. This yields the strongest single-solver performance (80.4\% Lite, 69.7\% Hard).

\textbf{Qwen SLR run 1: layered hypothesis generation with early exit.}
This solver parses predicates and argument domains, classifies predicates as car-level or train-level, and generates candidate body literals accordingly. It searches in ascending complexity layers, starting with one literal, then two, then three. Once a perfect rule is found within a layer, it exits early, thereby preferring the simplest correct rule. Candidates are purely conjunctive and are ranked by score and complexity.

\textbf{Qwen SLR run 2: in-Python filter + budget-limited verification.}
This solver moves most candidate filtering into Python to avoid expensive SWI-Prolog calls. It generates direct separating properties, arithmetic predicates, negation-as-failure rules, and universal negation rules, using both positive and negative examples. If no single-property rule passes, it escalates to two- and three-property conjunctions, but only sends a small top-$K$ set to the verifier. Despite the more elaborate design, performance is weaker: the verifier budget is too tight for medium and hard tasks, and the Python approximation may reject rules the official verifier would accept.

\paragraph{Claude Code SLR solver.}
The Claude Code SLR solver performs ascending-complexity search with a local Python evaluator. It parses facts into a normalized train-agnostic car model, re-indexing cars by \texttt{car\_num}, and dynamically discovers predicates. It then enumerates candidate rules by increasing complexity, including single-car, two-property, and two-car rule templates. The local evaluator emulates Prolog existential semantics and optionally re-scores top candidates with the official SWI-Prolog verifier. Its key strengths are train-id invariance, predicate discovery, and complexity-controlled search. It achieves 68.4\% overall accuracy, with 100\% on Basic, 78.4\% on Easy, and approximately 48\% on Medium/Hard; its high mean score indicates that failures are often near-misses rather than complete failures.

\paragraph{Summary pattern.}
CoT traces enable the agent to learn structural insights about the DSL (edit regions, feeding interactions, program ordering). Without CoT it defaults to brute-force beam search. With CoT but too few examples (12), it produces a competent but under-informed search. The best solvers share a common trait: an explicit mechanism for handling program-order interactions (permutation reorder, 2-step lookahead, or split-candidate pairs).

\paragraph{Key findings.}
First, CoT reasoning traces are critical for solver induction. Removing CoT from the 100-example demos drops performance from 53.4\% to 42.1\% on Lite and from 58.9\% to 24.8\% on Hard. Second, run-to-run variance is large: the three 100-example CoT Qwen runs span 53.4\%--79.2\% on Lite and 51.8\%--74.7\% on Hard, with each run inventing a qualitatively different algorithm. Third, 12 examples with CoT is insufficient, suggesting a lower bound on the amount of demonstration diversity needed. Fourth, the no-CoT solver takes more turns than the corresponding CoT run, suggesting that reasoning traces reduce the need for unguided probing. Finally, solver construction is cheap relative to inference, with single-run induction costs between \$0.30 and \$1.34.

\paragraph{Solver ensembles.}

\begin{table}[h]
\centering
\small
\caption{Symbolic solver ensembles.}
\smallskip
\label{tab:symbolic_solver_ablations}
\begin{tabular}{lccc}
\toprule
Ensemble & Lite Acc & Hard Acc & Tokens \\
\midrule
CC only & 80.4 & 69.7 & 0 \\
Best Qwen & 79.2 & 51.8 & 0 \\
3 Qwen (CoT) & 85.5 & 78.9 & 0 \\
All Qwen & 86.2 & 82.0 & 0 \\
All Symbolic & \textbf{91.3} & \textbf{84.7} & 0 \\
\bottomrule
\end{tabular}
\end{table}

Ensembling recovers variance across runs and yields the strongest overall performance, achieving 91.3\% on Lite and 84.7\% on Hard without any LLM inference cost.

\subsection{Additional Details: PBEBench-Hard}
\label{app:hard}

\paragraph{Search space.}
PBEBench-Hard uses the global alphabet ($V=52$: uppercase letters, lowercase letters, and digits) with cascades of length 2--20.
The per-program search space grows to $2.06 \times 10^{10}$ candidates; a length-20 cascade therefore has $\approx 1.8 \times 10^{206}$ possible program sequences.
At $10^9$ evaluations per second this would require $\approx 10^{190}$ years — far beyond any enumeration budget.
The induced symbolic solvers reduce this by extracting candidate predicates directly from input--output diffs, restricting replacement candidates to a small typed set, and applying beam or greedy search so that only hundreds to thousands of candidates are evaluated per task.

\begin{figure*}[!tbh]
    \centering
    \includegraphics[width=\textwidth]{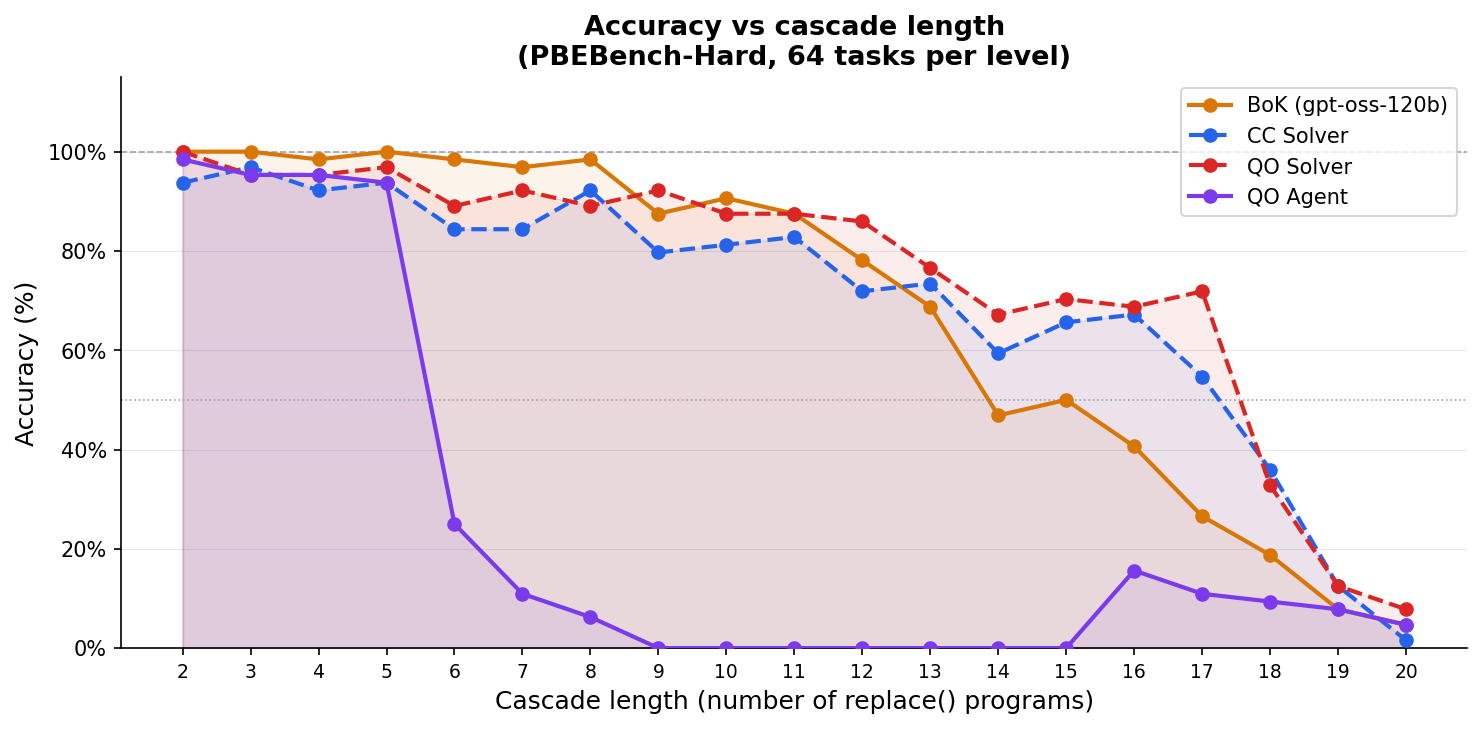}
    \caption{\textbf{PBEBench-Hard accuracy vs.\ cascade length.} Symbolic solvers (CC, QO) maintain substantially higher accuracy than BoK at long cascade lengths (CL~14+), where independent LLM sampling collapses.}
    \label{fig:pbebench_hard_accuracy}
\end{figure*}

\begin{figure*}[!tbh]
    \centering
    \includegraphics[width=\textwidth]{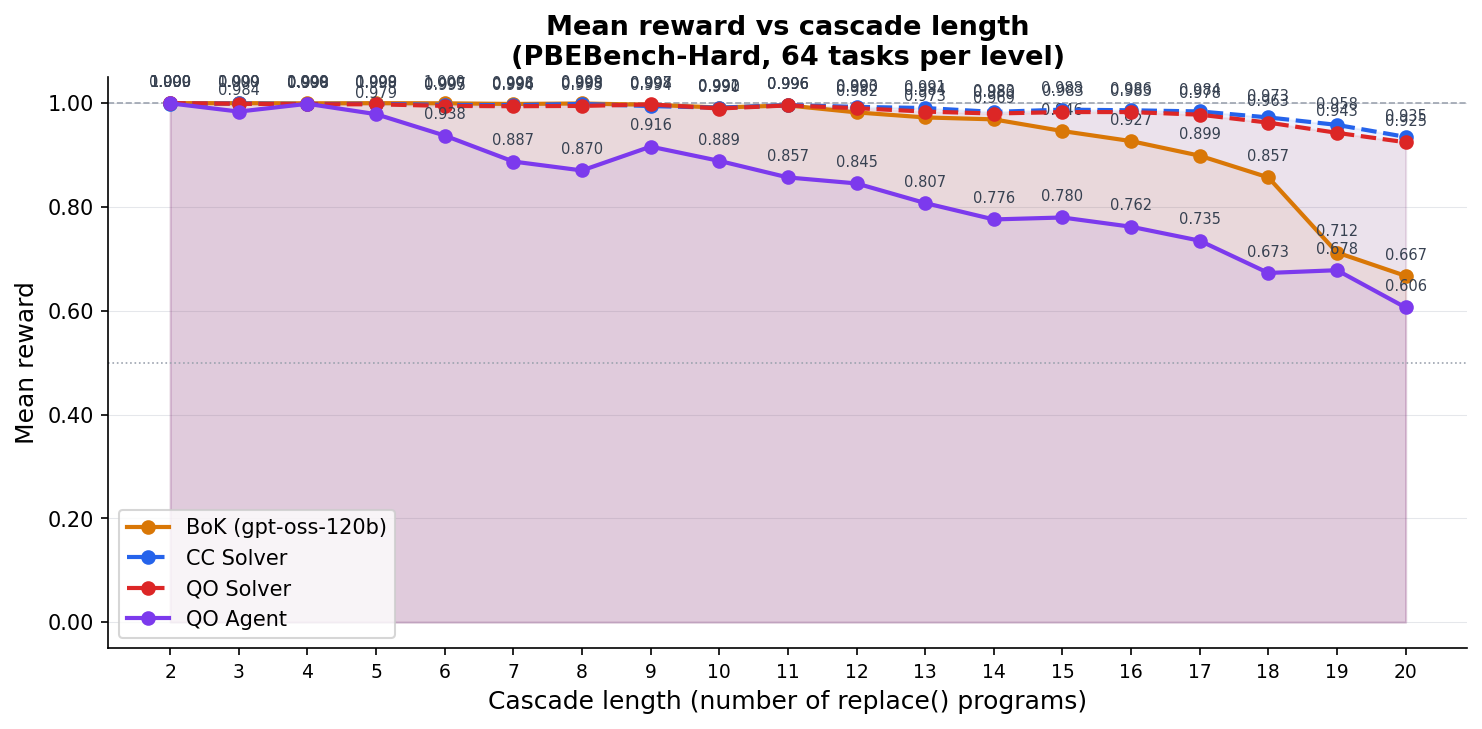}
    \caption{\textbf{PBEBench-Hard mean reward vs.\ cascade length.} Even at CL~20 where accuracy approaches zero, mean reward remains above 0.92 for both solvers, indicating near-misses rather than complete failures.}
    \label{fig:pbebench_hard_mean_reward}
\end{figure*}

\begin{figure*}[!tbh]
    \centering
    \includegraphics[width=\textwidth]{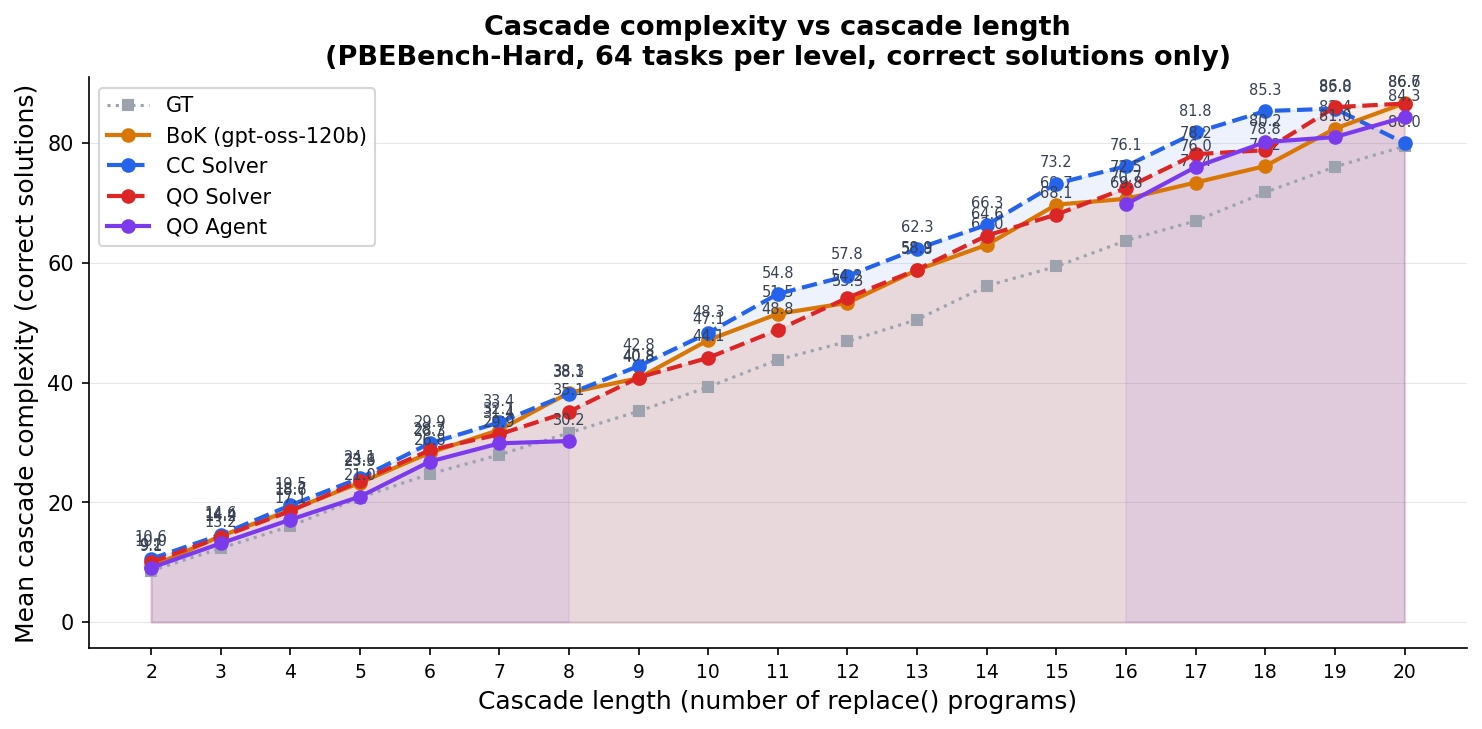}
    \caption{\textbf{PBEBench-Hard cascade complexity vs.\ cascade length.} Symbolic solvers overshoot ground-truth complexity substantially ($\Delta \approx +5$--$+8$) due to valid-but-verbose inductive solutions at long cascades; BoK complexity tracks GT more closely.}
    \label{fig:pbebench_hard_complexity}
\end{figure*}

\paragraph{Breakdown by cascade length.}
Table~\ref{tab:hard_by_cl} reports per-cascade-length accuracy and mean reward for both the CC and QO solvers.
Both solvers are strong at short cascades (CL 2--8, $\geq$84\%) and degrade gradually through CL 9--17, then collapse at CL 18--20 where even finding a partial near-miss requires a very long search horizon.
Despite near-zero accuracy at CL 20, mean reward remains above 0.92 at every level, confirming that failures are dominated by near-misses rather than complete failures.
Figures~\ref{fig:pbebench_hard_accuracy}, \ref{fig:pbebench_hard_mean_reward}, and \ref{fig:pbebench_hard_complexity} show the full accuracy, mean reward, and complexity profiles across cascade lengths, comparing CC and QO solvers with BoK and the QO direct coding-agent baseline.
In contrast to PBEBench-Lite, LLM test-time scaling (BoK) collapses at long cascades while the symbolic solvers maintain substantially higher accuracy through CL 17, illustrating the complementarity that makes hybrid inference most effective on Hard.

\begin{table}[t]
\centering
\caption{CC and QO solver accuracy and mean reward by cascade length on PBEBench-Hard (64 tasks per level).}
\smallskip
\label{tab:hard_by_cl}
\begin{tabular}{rrccccc}
\toprule
\textbf{CL} & \textbf{N} & \textbf{Acc\% (CC)} & \textbf{Reward (CC)} & \textbf{Acc\% (QO)} & \textbf{Reward (QO)} \\
\midrule
2  & 64 & 93.8 & 0.999 & 100.0 & 1.000 \\
3  & 64 & 96.9 & 0.999 &  95.3 & 0.999 \\
4  & 64 & 92.2 & 0.998 &  95.3 & 0.998 \\
5  & 64 & 93.8 & 0.999 &  96.9 & 0.998 \\
6  & 64 & 84.4 & 0.997 &  89.1 & 0.995 \\
7  & 64 & 84.4 & 0.996 &  92.2 & 0.994 \\
8  & 64 & 92.2 & 0.998 &  89.1 & 0.995 \\
9  & 64 & 79.7 & 0.994 &  92.2 & 0.998 \\
10 & 64 & 81.2 & 0.991 &  87.5 & 0.990 \\
11 & 64 & 82.8 & 0.996 &  87.5 & 0.996 \\
12 & 64 & 71.9 & 0.993 &  85.9 & 0.990 \\
13 & 64 & 73.4 & 0.991 &  76.6 & 0.984 \\
14 & 64 & 59.4 & 0.983 &  67.2 & 0.980 \\
15 & 64 & 65.6 & 0.988 &  70.3 & 0.983 \\
16 & 64 & 67.2 & 0.986 &  68.8 & 0.983 \\
17 & 64 & 54.7 & 0.984 &  71.9 & 0.978 \\
18 & 64 & 35.9 & 0.972 &  32.8 & 0.963 \\
19 & 64 & 12.5 & 0.958 &  12.5 & 0.943 \\
20 & 64 &  1.6 & 0.935 &   7.8 & 0.925 \\
\bottomrule
\end{tabular}
\end{table}

\paragraph{Breakdown by BFCC interaction type.}
Table~\ref{tab:hard_by_bfcc} reports CC solver accuracy across the 16 BFCC (bleeding, feeding, counterbleeding, counterfeeding) interaction categories.
Tasks with no BFCC interactions are easiest (92.7\%); adding feeding interactions consistently reduces accuracy more than bleeding alone.
The hardest category is all-four-BFCC (40.2\%), which constitutes 22\% of the dataset; dense mutual ordering interactions are the primary bottleneck for beam-search-based solvers.

\begin{table}[t]
\centering
\caption{CC solver accuracy by BFCC interaction type on PBEBench-Hard.}
\smallskip
\label{tab:hard_by_bfcc}
\begin{tabular}{lrr}
\toprule
\textbf{BFCC relationships} & \textbf{N} & \textbf{Acc\%} \\
\midrule
None                                                    &  55 & 92.7 \\
Bleeding                                                &  47 & 91.5 \\
Bleeding, Counterfeeding                                &  56 & 91.1 \\
Counterfeeding                                          &  64 & 90.6 \\
Counterbleeding                                         &  45 & 84.4 \\
Counterfeeding, Counterbleeding                         &  60 & 81.7 \\
Bleeding, Counterbleeding                               &  65 & 81.5 \\
Feeding                                                 &  64 & 75.0 \\
Feeding, Bleeding                                       &  59 & 78.0 \\
Feeding, Bleeding, Counterfeeding                       &  72 & 73.6 \\
Feeding, Counterfeeding, Counterbleeding                &  74 & 73.0 \\
Bleeding, Counterfeeding, Counterbleeding               &  77 & 71.4 \\
Feeding, Counterfeeding                                 &  76 & 65.8 \\
Feeding, Counterbleeding                                &  58 & 67.2 \\
Feeding, Bleeding, Counterbleeding                      &  78 & 66.7 \\
Feeding, Bleeding, Counterfeeding, Counterbleeding      & 266 & 40.2 \\
\bottomrule
\end{tabular}
\end{table}

\paragraph{Solver failure analysis.}
The dominant failure mode on both PBEBench-Lite and PBEBench-Hard is \emph{near-misses}: the solver finds a cascade that correctly transforms all but 1--3 input-output pairs, reflected in high mean rewards (0.944 on Lite, 0.987 on Hard) despite imperfect accuracy.
Three systematic causes account for most failures:
(1) tasks where the correct program can only be discovered through a long intermediate chain (cascade length $\geq 18$), where the beam horizon is exhausted before the necessary ordering is found;
(2) all-four-BFCC tasks that require holding multiple mutually interacting ordering constraints simultaneously within the beam;
(3) tasks where the pattern that fixes changed pairs also appears as a substring in unchanged pairs — the safe-phase beam discards it and the unrestricted phase does not recover in time.

\paragraph{Solver determinism.}
Both solvers were re-run on PBEBench-Hard under identical conditions to measure inference-time determinism.
Table~\ref{tab:hard_determinism} reports the results.
Note that the QO Solver row evaluates Run~1 (58.9\% Hard accuracy); the primary QO Solver result reported in the main paper uses Run~2 (74.7\%), which was selected based on best performance across induction runs.
Both runs are equivalent for measuring inference-time determinism, as we are measuring whether a fixed solver produces consistent outputs across independent evaluations.
All discrepancies between runs occur at the partial-credit boundary ($0.96 \leftrightarrow 1.0$ or $0.98 \leftrightarrow 1.0$); no task flips between clearly solved and clearly failed.
Both solvers are effectively deterministic: the CC solver agrees on 99.6\% of tasks across runs and the QO solver on 97.0\%.
Run-to-run variance in the paper therefore refers to independent \emph{induction} attempts (different solver algorithms), not to inference variance of a fixed solver.

\begin{table}[t]
\centering
\caption{Inference-time determinism of CC and QO solvers on PBEBench-Hard (two independent runs, identical settings).}
\smallskip
\label{tab:hard_determinism}
\begin{tabular}{lrrrr}
\toprule
\textbf{System} & \textbf{Run 1 Acc\%} & \textbf{Run 2 Acc\%} & \textbf{Agreement} & \textbf{Flips} \\
\midrule
CC Solver & 69.65 & 69.74 & 1211/1216 (99.6\%) & 5 \\
QO Solver & 58.88 & 57.89 & 1180/1216 (97.0\%) & 36 \\
\bottomrule
\end{tabular}
\end{table}

\subsection{Additional Details: Real-World Forward Reconstruction}
\label{app:real_fr}

We evaluate induced symbolic solvers on a real-world forward reconstruction dataset consisting of 3,077 language-pair tasks from Austronesian comparative linguistics. Each task requires inducing an ordered cascade of sound change rules from sparse input-output examples (median 4 examples). 
Compared to PBEBench, this setting introduces (i) an unseen IPA alphabet, (ii) variable and unknown cascade lengths, and (iii) no ground-truth programs.

\begin{table}[t]
\centering
\caption{Real-FR accuracy at different maximum cascade lengths ($k$). Rows marked $-$ were only evaluated at $k=100$. Run variance across three identical 100-example CoT induction runs (runs 1--3) is reported to illustrate algorithmic sensitivity.}
\smallskip
\label{tab:real_fr_main}
\begin{tabular}{lccc}
\toprule
System & $k=20$ & $k=50$ & $k=100$ \\
\midrule
\multicolumn{4}{l}{\textit{Individual solvers}} \\
CC Solver & 69.0 & 70.1 & \textbf{70.2} \\
Qwen Solver (run 2, 100ex+CoT) & 67.5 & 69.5 & \textbf{70.5} \\
Qwen Solver (run 1, 100ex+CoT) & --- & --- & 53.2 \\
Qwen Solver (run 3, 100ex+CoT) & --- & --- & 4.1$^\dagger$ \\
Qwen Solver (48ex+CoT) & --- & --- & 61.4 \\
Qwen Solver (12ex+CoT) & --- & --- & 29.4 \\
Qwen Solver (no-CoT) & --- & --- & 44.9 \\
\midrule
\multicolumn{4}{l}{\textit{Ensemble (union, $k=100$)}} \\
CC + Qwen run 2 & --- & --- & 76.1 \\
CC + Qwen runs 1+2 & --- & --- & 76.5 \\
All Qwen (5 solvers) & --- & --- & 78.9 \\
All solvers (excl.\ run 3) & --- & --- & \textbf{80.1} \\
\bottomrule
\end{tabular}
\smallskip\\
{\footnotesize $\dagger$ Run 3 uses a ``unique-op forcing'' strategy that fails under real IPA distributions.}
\end{table}

Both solvers achieve strong zero-shot performance (~70\%) without retraining. Increasing cascade length yields modest gains, after which performance saturates.

\paragraph{Effect of cascade length.}
Increasing the maximum number of programs (cascade length) from 20 to 100 yields modest gains (e.g., CC: 69.0\%→70.2\%, Qwen: 67.5\%→70.5\%), after which performance plateaus. 
This suggests that while real sound-change systems are deeper than PBEBench cascades, the primary bottleneck is candidate generation rather than search depth.

\paragraph{Solver variance and ensembling.}
Across independently induced solvers, performance ranges from 4.1\% to 70.5\%, reflecting substantial algorithmic variability. 
For example, three Qwen runs with identical training data achieve accuracies ranging from 4.1\% to 70.5\%. 
Some solver designs (e.g., unique-op forcing) fail under distribution shift, indicating sensitivity to assumptions about the training regime.
Combining solvers via union significantly improves performance, reaching up to 80.1\% accuracy, indicating that different solvers capture complementary strategies.

\paragraph{Compression analysis.}  To evaluate whether solvers discover compact rules or rely on memorization, we constrain cascade length relative to the number of examples. 
Under strong compression (programs limited to $\leq n_{\text{examples}}/5$), accuracy drops to $\approx$30\%, revealing a subset of tasks that admit compact, shared rules. 
The remaining $\approx$40 percentage points (from 30\% to 70\%) correspond to cases where solvers construct longer cascades that effectively patch individual examples rather than discovering general rules. Accuracy plateaus at ~30\% under strong compression, indicating that only a subset of tasks admit compact, shared rules with current solvers. The remaining gains arise from longer cascades that effectively patch individual examples.

\begin{table}[t]
\centering
\caption{Compression sweep for QO solver (run 2) on Real-FR. \texttt{max\_programs}~$= \max(2,\lceil n_{\text{examples}} / \text{ratio}\rceil)$.}
\label{tab:real_fr_compression}
\smallskip
\begin{tabular}{lcc}
\toprule
Setting & Acc\% & Avg programs \\
\midrule
Unconstrained ($k=100$) & 69.9 & 26.35 \\
ratio = 1.0 & 43.8 & 10.47 \\
ratio = 2.0 & 31.9 & 6.49 \\
ratio = 3.0 & 30.5 & 4.82 \\
ratio = 5.0 & \textbf{29.6} & \textbf{3.62} \\
\bottomrule
\end{tabular}
\end{table}

\paragraph{Sparse examples as the primary bottleneck.}
Real-FR tasks have a median of 4 examples per task (minimum 1, maximum 857), far sparser than PBEBench's guaranteed 5+ examples.
With 1--2 examples, multiple distinct cascades are consistent with the evidence, and any short program that happens to work scores 1.0 by coincidence.
Single-example tasks remain unsolvable in a principled sense regardless of compression ratio; the per-task accuracy gains from compression sweeps occur almost entirely on tasks with $\geq$3 examples.

\paragraph{Qualitative analysis.}
Under strong compression (ratio~$= 5$, programs $\leq \lceil n_\text{ex}/5 \rceil$), induced rules on multi-example tasks are often linguistically plausible sound laws, for example:
\begin{itemize}
    \item \texttt{replace('\#b', '\#f')} (Proto-Austronesian$\to$Amis): word-initial /b/→/f/ lenition — well-attested.
    \item \texttt{replace('}?\texttt{\#', 'h\#')} + \texttt{replace('wi', 'i')} ($\to$Bahasa Indonesia): word-final glottal→/h/ and /wi/→/i/ glide deletion — both documented Indonesian changes.
    \item \texttt{replace('\#}?\texttt{', '\#h')} + \texttt{replace('aw\#', 'au\#')} ($\to$Banjarese): word-initial /?/→/h/ and diphthong shift — classic Banjarese.
    \item \texttt{replace('}?\texttt{', '')} ($\to$Bidayuh): glottal stop deletion — extremely common cross-linguistically.
    \item \texttt{replace('da', 'ra')} + \texttt{replace('u}?\texttt{', 'o}?\texttt{')} ($\to$Atayal Squliq): /d/→/r/ and vowel lowering before glottal — both real Atayal changes.
\end{itemize}
In contrast, single-example tasks admit coincidental word-level substitutions regardless of compression ratio, and are noise by construction.

\paragraph{DSL expressivity vs.\ linguistic validity.}
The \texttt{replace(A,B)} DSL is strictly more expressive than real sound laws.
A genuine sound law maps one phoneme (or phoneme class in an environment) to another: /b/→/f/, /?/→$\emptyset$, /u/→/o/ before glottal.
The DSL permits arbitrary string substitutions with no phonological structure, e.g.\ \texttt{replace('m','ja')} would require a consonant to simultaneously become a vowel sequence — unknown in natural language.
The reward function (training accuracy) cannot distinguish these cases; both score 1.0. This has two consequences: (i)~false positives on sparse tasks ($\leq$2 examples), where almost any short program achieves 1.0 by coincidence; and (ii)~complexity inflation on rich tasks, where the unconstrained solver uses 26 programs on average — far more than any real sound-change history.

\paragraph{Reward function limitations and future directions.}
The current training-accuracy reward is appropriate for PBEBench (compact ground truth exists) but misaligned for Real-FR.
Four directions would improve evaluation: (1)~\emph{held-out evaluation} — even a 3/1 example split would filter coincidental single-word patches; (2)~\emph{complexity penalty} — e.g.\ $\text{reward} = \text{pass\_rate} - \lambda \cdot n_\text{programs}/n_\text{changed\_examples}$; (3)~\emph{phonological validity filter} — reject substitutions whose pattern spans more than one IPA segment; (4)~\emph{minimum example threshold} — exclude or down-weight tasks with fewer than 3--5 examples, which are noise-dominated regardless of method.

\subsection{Statistical Significance Testing}
\label{sec:appendix:statistical_testing}

\paragraph{Methodology.}
We test all pairwise comparisons between systems on PBEBench-Lite, PBEBench-Hard, and SLR-Bench using two complementary tests.
For binary per-task correctness (solved/unsolved), we use the \textbf{exact two-sided McNemar test}~\citep{mcnemar1947note}, which operates on discordant pairs — tasks where one system succeeds and the other fails — and is the standard choice for paired binary outcomes.
For continuous per-task mean reward, we use the \textbf{two-sided Wilcoxon signed-rank test}, a non-parametric paired test that makes no normality assumption and is appropriate for reward distributions that are heavy-tailed near 1.0.
Both tests are conducted at $\alpha = 0.05$; we report exact $p$-values with $^{*}p{<}0.05$, $^{**}p{<}0.01$, $^{***}p{<}0.001$.
No multiple-comparison correction is applied; these tests are exploratory and intended to distinguish genuine performance gaps from sampling noise.

The same tests are applied to SLR-Bench; results are in Table~\ref{tab:stat_tests_slr}.

\paragraph{PBEBench-Lite results (Tables~\ref{tab:stat_tests_lite_1}--\ref{tab:stat_tests_lite_3}).}
All comparisons between qualitatively distinct system categories (individual solvers vs.\ LLM scaling vs.\ hybrids) are strongly significant ($p{<}0.001$ by both tests), confirming that the accuracy differences are not due to chance.
Notable exceptions reveal the saturation regime: BoK vs.\ BoK+CC ($\Delta$Acc~$= +0.10$pp, $p = 1.0$) and BoK vs.\ BoK+All Sym ($\Delta$Acc $= +0.10$pp, $p = 1.0$) are not significant by either test, consistent with the ceiling effect.
The BoK vs.\ DF comparison ($\Delta$Acc $= -1.49$pp) reaches conventional significance by McNemar ($p = 0.020$) but not by Wilcoxon ($p = 0.595$), indicating that while BoK has a slight edge in binary solve rate, the two systems are essentially equivalent on partial-credit reward.
DF hybrids (DF+CC, DF+All Sym) show small but statistically significant improvements over DF alone ($p{<}0.01$), confirming that even marginal accuracy gains from solver augmentation are real on this dataset.
The accuracy difference between All Symbolic and BoK-based hybrids is significant ($p{<}0.001$), but the corresponding Wilcoxon $p$-values are not ($p > 0.15$), showing that hybrids close the accuracy gap through binary correctness on a handful of tasks without substantially changing the reward distribution.

\paragraph{PBEBench-Hard results (Table~\ref{tab:stat_tests_hard}).}
All comparisons are significant by both tests ($p{<}0.001$) with one exception: BoK vs.\ CC Solver ($\Delta$Acc $= +1.23$pp) is not significant by McNemar ($p = 0.379$), confirming that these two systems are statistically tied on accuracy despite the CC Solver's substantially higher mean reward ($\Delta$Rew $= +0.0445$, $p{<}0.001$).
This reflects the hard regime: BoK and CC Solver solve largely complementary subsets of tasks, so the overall pass rates are similar while per-task rewards diverge sharply.
Every hybrid improvement over its constituent components is significant ($p{<}0.001$), including the small BoK+All Sym vs.\ BoK+CC+QO difference ($+2.30$pp), validating that ensemble gains across the full solver suite are genuine.
Notably, BoK+CC vs.\ BoK+QO ($\Delta$Acc $= +1.23$pp, $p = 0.124$) is not significant, consistent with the near-equivalent performance of the two individual solvers on this split.


\begin{table}[!tbh]
\centering
\small
\caption{PBEBench-Lite pairwise significance tests (part 1/3): LLM baselines as System~A.
$\Delta$Acc = B$-$A accuracy (pp); $\Delta$Rew = B$-$A mean reward.
\textit{p}(Mc): exact two-sided McNemar test. \textit{p}(Wi): two-sided Wilcoxon signed-rank test.
$^{*}p{<}0.05$, $^{**}p{<}0.01$, $^{***}p{<}0.001$.}
\smallskip
\label{tab:stat_tests_lite_1}

\begin{tabular}{llrrrr}
\toprule
\textbf{System A} & \textbf{System B} & $\boldsymbol{\Delta}$\textbf{Acc} & \textit{p}(Mc) & $\boldsymbol{\Delta}$\textbf{Rew} & \textit{p}(Wi) \\
\midrule
BoK & DF                & $-$1.49 & 0.020$^{*}$      & $-$0.0012 & 0.595 \\
BoK & QO Agent          & $-$6.65 & $<\!0.001^{***}$ & $-$0.0264 & $<\!0.001^{***}$ \\
BoK & CC Solver         & $-$13.49 & $<\!0.001^{***}$ & $-$0.0369 & $<\!0.001^{***}$ \\
BoK & QO Solver         & $-$28.17 & $<\!0.001^{***}$ & $-$0.0786 & $<\!0.001^{***}$ \\
BoK & All Symbolic      & $-$2.58 & 0.001$^{**}$     & $-$0.0036 & 0.303 \\
BoK & BoK + QO          & $+$0.00 & 1.000            & $+$0.0000 & 1.000 \\
BoK & BoK + CC          & $+$0.10 & 1.000            & $+$0.0002 & 0.317 \\
BoK & BoK + All Sym     & $+$0.10 & 1.000            & $+$0.0002 & 0.317 \\
BoK & DF + QO           & $-$0.99 & 0.110            & $+$0.0002 & 0.976 \\
BoK & DF + CC           & $-$0.79 & 0.200            & $+$0.0008 & 0.729 \\
BoK & DF + All Sym      & $-$0.69 & 0.265            & $+$0.0010 & 0.661 \\
\midrule
DF  & QO Agent          & $-$5.16 & $<\!0.001^{***}$ & $-$0.0252 & $<\!0.001^{***}$ \\
DF  & CC Solver         & $-$12.00 & $<\!0.001^{***}$ & $-$0.0357 & $<\!0.001^{***}$ \\
DF  & QO Solver         & $-$26.69 & $<\!0.001^{***}$ & $-$0.0774 & $<\!0.001^{***}$ \\
DF  & All Symbolic      & $-$1.09 & 0.235            & $-$0.0024 & 0.580 \\
DF  & BoK + QO          & $+$1.49 & 0.020$^{*}$      & $+$0.0012 & 0.595 \\
DF  & BoK + CC          & $+$1.59 & 0.011$^{*}$      & $+$0.0014 & 0.540 \\
DF  & BoK + All Sym     & $+$1.59 & 0.011$^{*}$      & $+$0.0014 & 0.540 \\
DF  & DF + QO           & $+$0.50 & 0.062            & $+$0.0014 & 0.034$^{*}$ \\
DF  & DF + CC           & $+$0.69 & 0.016$^{*}$      & $+$0.0020 & 0.014$^{*}$ \\
DF  & DF + All Sym      & $+$0.79 & 0.008$^{**}$     & $+$0.0022 & 0.008$^{**}$ \\
\bottomrule
\end{tabular}%
\end{table}

\begin{table}[!tbh]
\centering
\small
\caption{PBEBench-Lite pairwise significance tests (part 2/3): coding-agent baseline and individual symbolic solvers as System~A. Column headers as in part~1.}
\smallskip
\label{tab:stat_tests_lite_2}
\begin{tabular}{llrrrr}
\toprule
\textbf{System A} & \textbf{System B} & $\boldsymbol{\Delta}$\textbf{Acc} & \textit{p}(Mc) & $\boldsymbol{\Delta}$\textbf{Rew} & \textit{p}(Wi) \\
\midrule
QO Agent & CC Solver      & $-$6.85  & $<\!0.001^{***}$ & $-$0.0105 & 0.064 \\
QO Agent & QO Solver      & $-$21.53 & $<\!0.001^{***}$ & $-$0.0522 & $<\!0.001^{***}$ \\
QO Agent & All Symbolic   & $+$4.07  & $<\!0.001^{***}$ & $+$0.0228 & $<\!0.001^{***}$ \\
QO Agent & BoK + QO       & $+$6.65  & $<\!0.001^{***}$ & $+$0.0264 & $<\!0.001^{***}$ \\
QO Agent & BoK + CC       & $+$6.75  & $<\!0.001^{***}$ & $+$0.0266 & $<\!0.001^{***}$ \\
QO Agent & BoK + All Sym  & $+$6.75  & $<\!0.001^{***}$ & $+$0.0266 & $<\!0.001^{***}$ \\
QO Agent & DF + QO        & $+$5.65  & $<\!0.001^{***}$ & $+$0.0266 & $<\!0.001^{***}$ \\
QO Agent & DF + CC        & $+$5.85  & $<\!0.001^{***}$ & $+$0.0272 & $<\!0.001^{***}$ \\
QO Agent & DF + All Sym   & $+$5.95  & $<\!0.001^{***}$ & $+$0.0274 & $<\!0.001^{***}$ \\
\midrule
CC Solver & QO Solver     & $-$14.68 & $<\!0.001^{***}$ & $-$0.0417 & $<\!0.001^{***}$ \\
CC Solver & All Symbolic  & $+$10.91 & $<\!0.001^{***}$ & $+$0.0333 & $<\!0.001^{***}$ \\
CC Solver & BoK + QO      & $+$13.49 & $<\!0.001^{***}$ & $+$0.0369 & $<\!0.001^{***}$ \\
CC Solver & BoK + CC      & $+$13.59 & $<\!0.001^{***}$ & $+$0.0371 & $<\!0.001^{***}$ \\
CC Solver & BoK + All Sym & $+$13.59 & $<\!0.001^{***}$ & $+$0.0371 & $<\!0.001^{***}$ \\
CC Solver & DF + QO       & $+$12.50 & $<\!0.001^{***}$ & $+$0.0371 & $<\!0.001^{***}$ \\
CC Solver & DF + CC       & $+$12.70 & $<\!0.001^{***}$ & $+$0.0377 & $<\!0.001^{***}$ \\
CC Solver & DF + All Sym  & $+$12.80 & $<\!0.001^{***}$ & $+$0.0379 & $<\!0.001^{***}$ \\
\midrule
QO Solver & All Symbolic  & $+$25.60 & $<\!0.001^{***}$ & $+$0.0750 & $<\!0.001^{***}$ \\
QO Solver & BoK + QO      & $+$28.17 & $<\!0.001^{***}$ & $+$0.0786 & $<\!0.001^{***}$ \\
QO Solver & BoK + CC      & $+$28.27 & $<\!0.001^{***}$ & $+$0.0788 & $<\!0.001^{***}$ \\
QO Solver & BoK + All Sym & $+$28.27 & $<\!0.001^{***}$ & $+$0.0788 & $<\!0.001^{***}$ \\
QO Solver & DF + QO       & $+$27.18 & $<\!0.001^{***}$ & $+$0.0788 & $<\!0.001^{***}$ \\
QO Solver & DF + CC       & $+$27.38 & $<\!0.001^{***}$ & $+$0.0794 & $<\!0.001^{***}$ \\
QO Solver & DF + All Sym  & $+$27.48 & $<\!0.001^{***}$ & $+$0.0796 & $<\!0.001^{***}$ \\
\bottomrule
\end{tabular}
\end{table}

\begin{table}[!tbh]
\centering
\small
\caption{PBEBench-Lite pairwise significance tests (part 3/3): All Symbolic and hybrid-vs-hybrid comparisons. Column headers as in part~1.}
\smallskip
\label{tab:stat_tests_lite_3}
\begin{tabular}{llrrrr}
\toprule
\textbf{System A} & \textbf{System B} & $\boldsymbol{\Delta}$\textbf{Acc} & \textit{p}(Mc) & $\boldsymbol{\Delta}$\textbf{Rew} & \textit{p}(Wi) \\
\midrule
All Symbolic & BoK + QO      & $+$2.58 & 0.001$^{**}$     & $+$0.0036 & 0.303 \\
All Symbolic & BoK + CC      & $+$2.68 & $<\!0.001^{***}$ & $+$0.0038 & 0.271 \\
All Symbolic & BoK + All Sym & $+$2.68 & $<\!0.001^{***}$ & $+$0.0038 & 0.271 \\
All Symbolic & DF + QO       & $+$1.59 & 0.064            & $+$0.0038 & 0.289 \\
All Symbolic & DF + CC       & $+$1.79 & 0.033$^{*}$      & $+$0.0044 & 0.179 \\
All Symbolic & DF + All Sym  & $+$1.88 & 0.023$^{*}$      & $+$0.0046 & 0.155 \\
\midrule
BoK + QO  & BoK + CC      & $+$0.10 & 1.000 & $+$0.0002 & 0.317 \\
BoK + QO  & BoK + All Sym & $+$0.10 & 1.000 & $+$0.0002 & 0.317 \\
BoK + QO  & DF + QO       & $-$0.99 & 0.110 & $+$0.0002 & 0.976 \\
BoK + QO  & DF + CC       & $-$0.79 & 0.200 & $+$0.0008 & 0.729 \\
BoK + QO  & DF + All Sym  & $-$0.69 & 0.265 & $+$0.0010 & 0.661 \\
\midrule
BoK + CC  & BoK + All Sym & $+$0.00 & 1.000 & $+$0.0000 & 1.000 \\
BoK + CC  & DF + QO       & $-$1.09 & 0.071 & $+$0.0000 & 0.959 \\
BoK + CC  & DF + CC       & $-$0.89 & 0.136 & $+$0.0006 & 0.792 \\
BoK + CC  & DF + All Sym  & $-$0.79 & 0.185 & $+$0.0008 & 0.721 \\
\midrule
BoK + All Sym & DF + QO      & $-$1.09 & 0.071 & $+$0.0000 & 0.959 \\
BoK + All Sym & DF + CC      & $-$0.89 & 0.136 & $+$0.0006 & 0.792 \\
BoK + All Sym & DF + All Sym & $-$0.79 & 0.185 & $+$0.0008 & 0.721 \\
\midrule
DF + QO  & DF + CC      & $+$0.20 & 0.625 & $+$0.0006 & 0.257 \\
DF + QO  & DF + All Sym & $+$0.30 & 0.250 & $+$0.0008 & 0.102 \\
DF + CC  & DF + All Sym & $+$0.10 & 1.000 & $+$0.0002 & 0.317 \\
\bottomrule
\end{tabular}%
\end{table}

\begin{table*}[!tbh]
\centering
\caption{Pairwise significance tests on PBEBench-Hard. $\Delta$Acc = B$-$A accuracy difference (pp). $\Delta$Rew = B$-$A mean reward difference. \textit{p}(McNemar) uses the exact two-sided McNemar test on per-task binary correctness. \textit{p}(Wilcoxon) uses the two-sided Wilcoxon signed-rank test on per-task rewards. $^{*}p<0.05$, $^{**}p<0.01$, $^{***}p<0.001$.}
\smallskip
\label{tab:stat_tests_hard}
\resizebox{\textwidth}{!}{%
\begin{tabular}{llrrrr}
\toprule
\textbf{System A} & \textbf{System B} & $\boldsymbol{\Delta}$\textbf{Acc\%} & \textit{p}(McNemar) & $\boldsymbol{\Delta}$\textbf{Rew} & \textit{p}(Wilcoxon) \\
\midrule
BoK & CC Solver & +1.23 & 0.379 & +0.0445 & $<\!0.001^{***}$ \\
BoK & QO Solver & +6.25 & $<\!0.001^{***}$ & +0.0408 & $<\!0.001^{***}$ \\
BoK & All Symbolic & +16.28 & $<\!0.001^{***}$ & +0.0493 & $<\!0.001^{***}$ \\
BoK & BoK + CC & +11.02 & $<\!0.001^{***}$ & +0.0080 & $<\!0.001^{***}$ \\
BoK & BoK + QO & +12.25 & $<\!0.001^{***}$ & +0.0069 & $<\!0.001^{***}$ \\
BoK & BoK + CC + QO & +15.05 & $<\!0.001^{***}$ & +0.0103 & $<\!0.001^{***}$ \\
BoK & BoK + All Sym & +17.35 & $<\!0.001^{***}$ & +0.0142 & $<\!0.001^{***}$ \\
CC Solver & QO Solver & +5.02 & $<\!0.001^{***}$ & -0.0037 & $<\!0.001^{***}$ \\
CC Solver & All Symbolic & +15.05 & $<\!0.001^{***}$ & +0.0048 & $<\!0.001^{***}$ \\
CC Solver & BoK + CC & +9.79 & $<\!0.001^{***}$ & -0.0365 & $<\!0.001^{***}$ \\
CC Solver & BoK + QO & +11.02 & $<\!0.001^{***}$ & -0.0376 & $<\!0.001^{***}$ \\
CC Solver & BoK + CC + QO & +13.82 & $<\!0.001^{***}$ & -0.0342 & $<\!0.001^{***}$ \\
CC Solver & BoK + All Sym & +16.12 & $<\!0.001^{***}$ & -0.0303 & $<\!0.001^{***}$ \\
QO Solver & All Symbolic & +10.03 & $<\!0.001^{***}$ & +0.0085 & $<\!0.001^{***}$ \\
QO Solver & BoK + CC & +4.77 & $<\!0.001^{***}$ & -0.0328 & $<\!0.001^{***}$ \\
QO Solver & BoK + QO & +6.00 & $<\!0.001^{***}$ & -0.0339 & $<\!0.001^{***}$ \\
QO Solver & BoK + CC + QO & +8.80 & $<\!0.001^{***}$ & -0.0305 & $<\!0.001^{***}$ \\
QO Solver & BoK + All Sym & +11.10 & $<\!0.001^{***}$ & -0.0266 & $<\!0.001^{***}$ \\
All Symbolic & BoK + CC & -5.26 & $<\!0.001^{***}$ & -0.0412 & $<\!0.001^{***}$ \\
All Symbolic & BoK + QO & -4.03 & $<\!0.001^{***}$ & -0.0424 & $<\!0.001^{***}$ \\
All Symbolic & BoK + CC + QO & -1.23 & 0.028$^{*}$ & -0.0390 & $<\!0.001^{***}$ \\
All Symbolic & BoK + All Sym & +1.07 & $<\!0.001^{***}$ & -0.0350 & $<\!0.001^{***}$ \\
BoK + CC & BoK + QO & +1.23 & 0.124 & -0.0012 & 0.292 \\
BoK + CC & BoK + CC + QO & +4.03 & $<\!0.001^{***}$ & +0.0022 & $<\!0.001^{***}$ \\
BoK + CC & BoK + All Sym & +6.33 & $<\!0.001^{***}$ & +0.0062 & $<\!0.001^{***}$ \\
BoK + QO & BoK + CC + QO & +2.80 & $<\!0.001^{***}$ & +0.0034 & $<\!0.001^{***}$ \\
BoK + QO & BoK + All Sym & +5.10 & $<\!0.001^{***}$ & +0.0074 & $<\!0.001^{***}$ \\
BoK + CC + QO & BoK + All Sym & +2.30 & $<\!0.001^{***}$ & +0.0040 & $<\!0.001^{***}$ \\
\bottomrule
\end{tabular}%
}
\end{table*}

\paragraph{SLR-Bench results (Table~\ref{tab:stat_tests_slr}).}
Nearly all comparisons are significant by both tests ($p{<}0.001$), with a few interpretable exceptions.
BoK vs.\ CC Solver ($\Delta$Acc $= -0.30$pp, $p = 0.896$) is not significant by McNemar, confirming these two systems are statistically tied in binary accuracy despite CC Solver having substantially higher mean reward ($\Delta$Rew $= +0.027$, $p{<}0.001$) — consistent with the finding that CC Solver produces fewer complete failures but near-identical solve rates.
DF vs.\ BoK+CC ($\Delta$Acc $= +0.70$pp, $p = 0.621$) is also not significant, indicating that simply adding the CC Solver to BoK matches DF's accuracy without a statistically distinguishable difference.
In contrast, every DF-based hybrid provides a strongly significant improvement over DF alone ($p{<}0.001$), confirming that symbolic solvers add genuine complementary coverage beyond what DF's sequential search achieves.
The BoK+QO vs.\ DF+QO comparison ($\Delta$Acc $= +7.80$pp, $p{<}0.001$) is significant by McNemar but not by Wilcoxon ($p = 0.359$), reflecting that DF+QO's accuracy gain comes from resolving tasks at the binary solve boundary without changing the reward distribution on partial-credit tasks.

\begin{table*}[!tbh]
\centering
\caption{Pairwise significance tests on SLR-Bench. $\Delta$Acc = B$-$A accuracy difference (pp). $\Delta$Rew = B$-$A mean reward difference. \textit{p}(McNemar) uses the exact two-sided McNemar test on per-task binary correctness. \textit{p}(Wilcoxon) uses the two-sided Wilcoxon signed-rank test on per-task rewards. $^{*}p{<}0.05$, $^{**}p{<}0.01$, $^{***}p{<}0.001$.}
\smallskip
\label{tab:stat_tests_slr}
\resizebox{\textwidth}{!}{%
\begin{tabular}{llrrrr}
\toprule
\textbf{System A} & \textbf{System B} & $\boldsymbol{\Delta}$\textbf{Acc\%} & \textit{p}(McNemar) & $\boldsymbol{\Delta}$\textbf{Rew} & \textit{p}(Wilcoxon) \\
\midrule
BoK & DF & +10.90 & $<\!0.001^{***}$ & +0.0086 & $<\!0.001^{***}$ \\
BoK & CC Solver & $-$0.30 & 0.896 & +0.0265 & $<\!0.001^{***}$ \\
BoK & QO Solver & $-$8.00 & $<\!0.001^{***}$ & $-$0.3334 & $<\!0.001^{***}$ \\
BoK & BoK + QO & +7.20 & $<\!0.001^{***}$ & +0.0141 & $<\!0.001^{***}$ \\
BoK & BoK + CC & +11.60 & $<\!0.001^{***}$ & +0.0215 & $<\!0.001^{***}$ \\
BoK & BoK + CC + QO & +11.60 & $<\!0.001^{***}$ & +0.0215 & $<\!0.001^{***}$ \\
BoK & DF + QO & +15.00 & $<\!0.001^{***}$ & $-$0.0054 & 0.005$^{**}$ \\
BoK & DF + CC & +17.80 & $<\!0.001^{***}$ & +0.0316 & $<\!0.001^{***}$ \\
BoK & DF + CC + QO & +17.90 & $<\!0.001^{***}$ & +0.0317 & $<\!0.001^{***}$ \\
\midrule
DF & CC Solver & $-$11.20 & $<\!0.001^{***}$ & +0.0179 & 0.003$^{**}$ \\
DF & QO Solver & $-$18.90 & $<\!0.001^{***}$ & $-$0.3420 & $<\!0.001^{***}$ \\
DF & BoK + QO & $-$3.70 & 0.002$^{**}$ & +0.0054 & 0.564 \\
DF & BoK + CC & +0.70 & 0.621 & +0.0129 & 0.006$^{**}$ \\
DF & BoK + CC + QO & +0.70 & 0.621 & +0.0129 & 0.006$^{**}$ \\
DF & DF + QO & +4.10 & $<\!0.001^{***}$ & $-$0.0140 & 0.005$^{**}$ \\
DF & DF + CC & +6.90 & $<\!0.001^{***}$ & +0.0230 & $<\!0.001^{***}$ \\
DF & DF + CC + QO & +7.00 & $<\!0.001^{***}$ & +0.0230 & $<\!0.001^{***}$ \\
\midrule
CC Solver & QO Solver & $-$7.70 & $<\!0.001^{***}$ & $-$0.3599 & $<\!0.001^{***}$ \\
CC Solver & BoK + QO & +7.50 & $<\!0.001^{***}$ & $-$0.0124 & $<\!0.001^{***}$ \\
CC Solver & BoK + CC & +11.90 & $<\!0.001^{***}$ & $-$0.0050 & 0.131 \\
CC Solver & BoK + CC + QO & +11.90 & $<\!0.001^{***}$ & $-$0.0050 & 0.131 \\
CC Solver & DF + QO & +15.30 & $<\!0.001^{***}$ & $-$0.0319 & 0.123 \\
CC Solver & DF + CC & +18.10 & $<\!0.001^{***}$ & +0.0051 & $<\!0.001^{***}$ \\
CC Solver & DF + CC + QO & +18.20 & $<\!0.001^{***}$ & +0.0052 & $<\!0.001^{***}$ \\
\midrule
QO Solver & BoK + QO & +15.20 & $<\!0.001^{***}$ & +0.3474 & $<\!0.001^{***}$ \\
QO Solver & BoK + CC & +19.60 & $<\!0.001^{***}$ & +0.3549 & $<\!0.001^{***}$ \\
QO Solver & BoK + CC + QO & +19.60 & $<\!0.001^{***}$ & +0.3549 & $<\!0.001^{***}$ \\
QO Solver & DF + QO & +23.00 & $<\!0.001^{***}$ & +0.3280 & $<\!0.001^{***}$ \\
QO Solver & DF + CC & +25.80 & $<\!0.001^{***}$ & +0.3650 & $<\!0.001^{***}$ \\
QO Solver & DF + CC + QO & +25.90 & $<\!0.001^{***}$ & +0.3650 & $<\!0.001^{***}$ \\
\midrule
BoK + QO & BoK + CC & +4.40 & $<\!0.001^{***}$ & +0.0074 & $<\!0.001^{***}$ \\
BoK + QO & BoK + CC + QO & +4.40 & $<\!0.001^{***}$ & +0.0074 & $<\!0.001^{***}$ \\
BoK + QO & DF + QO & +7.80 & $<\!0.001^{***}$ & $-$0.0194 & 0.359 \\
BoK + QO & DF + CC & +10.60 & $<\!0.001^{***}$ & +0.0176 & $<\!0.001^{***}$ \\
BoK + QO & DF + CC + QO & +10.70 & $<\!0.001^{***}$ & +0.0176 & $<\!0.001^{***}$ \\
\midrule
BoK + CC & BoK + CC + QO & +0.00 & 1.000 & +0.0000 & 1.000 \\
BoK + CC & DF + QO & +3.40 & 0.001$^{**}$ & $-$0.0269 & 0.008$^{**}$ \\
BoK + CC & DF + CC & +6.20 & $<\!0.001^{***}$ & +0.0101 & $<\!0.001^{***}$ \\
BoK + CC & DF + CC + QO & +6.30 & $<\!0.001^{***}$ & +0.0102 & $<\!0.001^{***}$ \\
\midrule
BoK + CC + QO & DF + QO & +3.40 & 0.001$^{**}$ & $-$0.0269 & 0.008$^{**}$ \\
BoK + CC + QO & DF + CC & +6.20 & $<\!0.001^{***}$ & +0.0101 & $<\!0.001^{***}$ \\
BoK + CC + QO & DF + CC + QO & +6.30 & $<\!0.001^{***}$ & +0.0102 & $<\!0.001^{***}$ \\
\midrule
DF + QO & DF + CC & +2.80 & $<\!0.001^{***}$ & +0.0370 & $<\!0.001^{***}$ \\
DF + QO & DF + CC + QO & +2.90 & $<\!0.001^{***}$ & +0.0370 & $<\!0.001^{***}$ \\
DF + CC & DF + CC + QO & +0.10 & 1.000 & +0.0000 & 0.317 \\
\bottomrule
\end{tabular}%
}
\end{table*}

\subsection{Additional Details: SLR-Bench}
\label{app:slr}

\paragraph{Search space.}
SLR-Bench requires inducing a Prolog rule \texttt{eastbound(T) :- has\_car(T,C), lit$_1$, \ldots} with up to 4 body literals drawn from $L$ ground literals per task.
The candidate count is $\binom{L}{1}+\binom{L}{2}+\binom{L}{3}+\binom{L}{4}$, where $L$ grows with curriculum level.
Each candidate requires a SWI-Prolog subprocess call (50--200\,ms), making brute force infeasible at levels 15--20.
The induced solvers address this by (1) extracting only task-relevant literals, (2) searching layer-by-layer and stopping at the first layer with a perfect rule, and (3) pruning by syntactic features before evaluation.

\begin{table}[!tbh]
\centering
\caption{SLR-Bench search space by curriculum level.}
\smallskip
\resizebox{\columnwidth}{!}{%
\begin{tabular}{rrrrr}
\toprule
\textbf{Curriculum level} & \textbf{Ground literals $L$} & \textbf{Candidates ($\leq$4-body)} & \textbf{@50\,ms/eval} & \textbf{@200\,ms/eval} \\
\midrule
1  &  5 &         30 & ${<}1$\,s  & ${<}1$\,s  \\
5  & 11 &        561 & ${<}1$\,s  & ${<}1$\,s  \\
10 & 19 &      5,035 & ${\sim}4$\,min & ${\sim}17$\,min \\
15 & 33 &     46,937 & ${\sim}39$\,min & ${\sim}2.6$\,hr \\
20 & 50 &    251,175 & ${\sim}3.5$\,hr & ${\sim}14$\,hr \\
\bottomrule
\end{tabular}%
}
\label{tab:slr_search_space}
\end{table}

\begin{figure*}[!tbh]
\centering
\begin{subfigure}{0.48\textwidth}
    \centering
    \includegraphics[width=\linewidth]{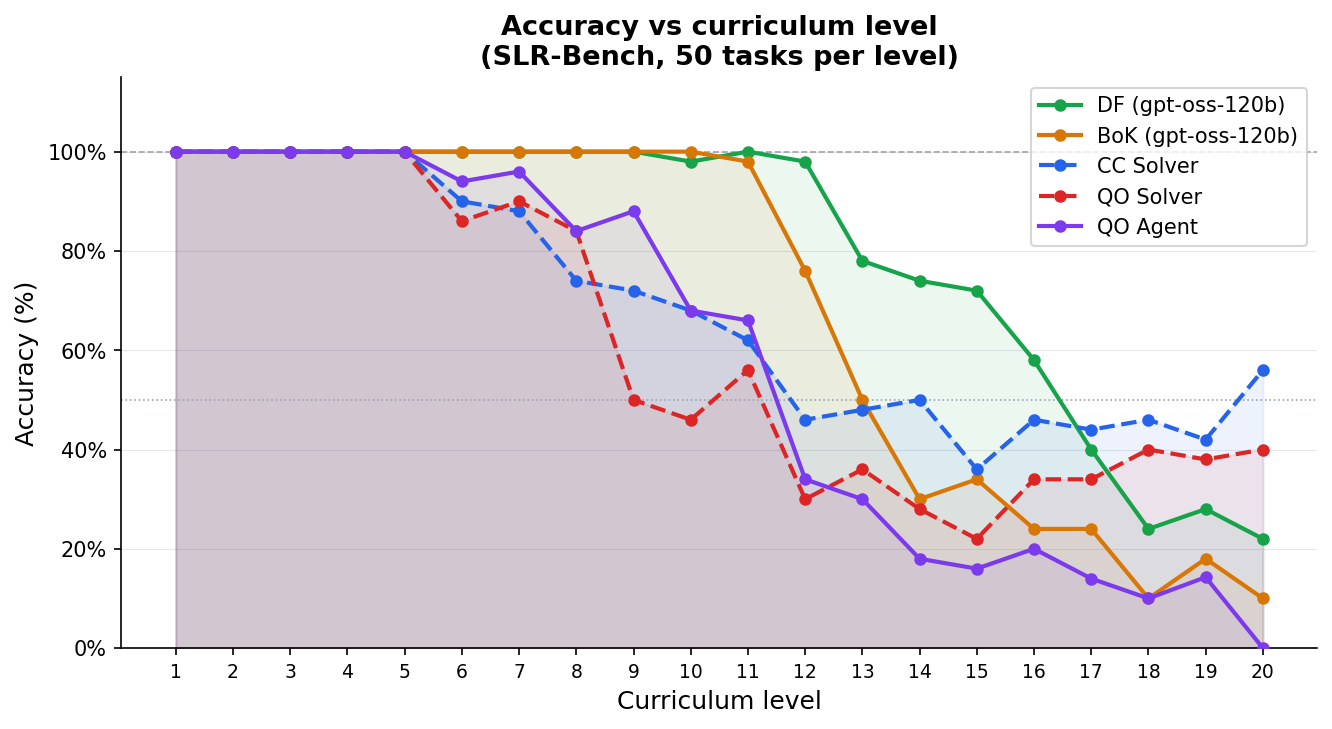}
    \caption{Pass rate by curriculum level.}
    \label{fig:slr_level_passrate}
\end{subfigure}\hfill
\begin{subfigure}{0.48\textwidth}
    \centering
    \includegraphics[width=\linewidth]{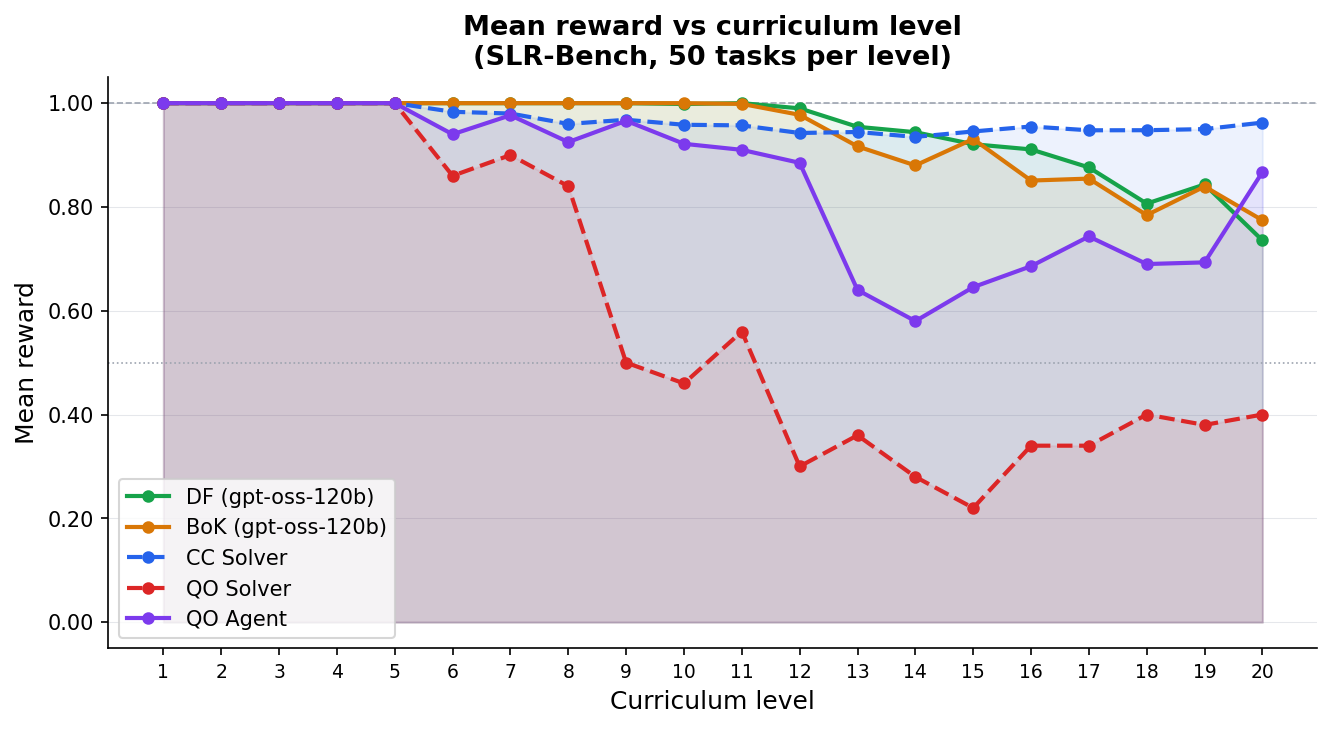}
    \caption{Mean reward by curriculum level.}
    \label{fig:slr_level_reward}
\end{subfigure}
\caption{\textbf{SLR-Bench: per-level breakdown.} LLM methods dominate at low curriculum levels; symbolic solvers sustain accuracy at Hard levels (16--20) where LLMs collapse.}
\label{fig:slr_level_breakdown}
\end{figure*}

\paragraph{Per-tier and per-level accuracy.}
Figure~\ref{fig:slr_level_passrate} shows pass rate by curriculum level for all systems.
LLM methods (BoK, DF) are strong through levels 1--10 but degrade sharply from level 11 onward, with BoK collapsing to near-zero at levels 16--20.
In contrast, the CC Solver maintains 30--50\% accuracy at levels 16--20 and achieves the highest Hard-tier pass rate of any single system.

\paragraph{Symbolic solver breakdown by rule complexity.}
Complexity groupings follow the binning used in the original SLR-Bench paper~\citep{helff2025slr}.
Table~\ref{tab:slr_solver_by_complexity} shows that CC Solver accuracy degrades gracefully with rule complexity, maintaining mean reward above 0.94 at all complexity levels.
In contrast, QO Solver's mean reward collapses at complexity 2+ (0.32--0.51), reflecting a larger fraction of complete failures rather than near-misses.

\begin{table}[!tbh]
\centering
\caption{SLR-Bench symbolic solver accuracy by ground-truth rule complexity.}
\smallskip
\resizebox{\columnwidth}{!}{%
\begin{tabular}{rrrrrr}
\toprule
\textbf{Complexity} & \textbf{N} & \textbf{Acc\% (CC)} & \textbf{Reward (CC)} & \textbf{Acc\% (QO)} & \textbf{Reward (QO)} \\
\midrule
1     &  50 & 100.0\% & 1.0000 & 100.0\% & 1.0000 \\
1--2  & 350 &  93.1\% & 0.9890 &  94.3\% & 0.9429 \\
2--3  & 150 &  67.3\% & 0.9612 &  50.7\% & 0.5067 \\
3--4  & 100 &  47.0\% & 0.9435 &  33.0\% & 0.3300 \\
4--5  & 250 &  44.4\% & 0.9462 &  31.6\% & 0.3160 \\
5     & 100 &  49.0\% & 0.9563 &  39.0\% & 0.3900 \\
\bottomrule
\end{tabular}%
}
\label{tab:slr_solver_by_complexity}
\end{table}

\begin{figure}[!tbh]
\centering
\includegraphics[width=\columnwidth]{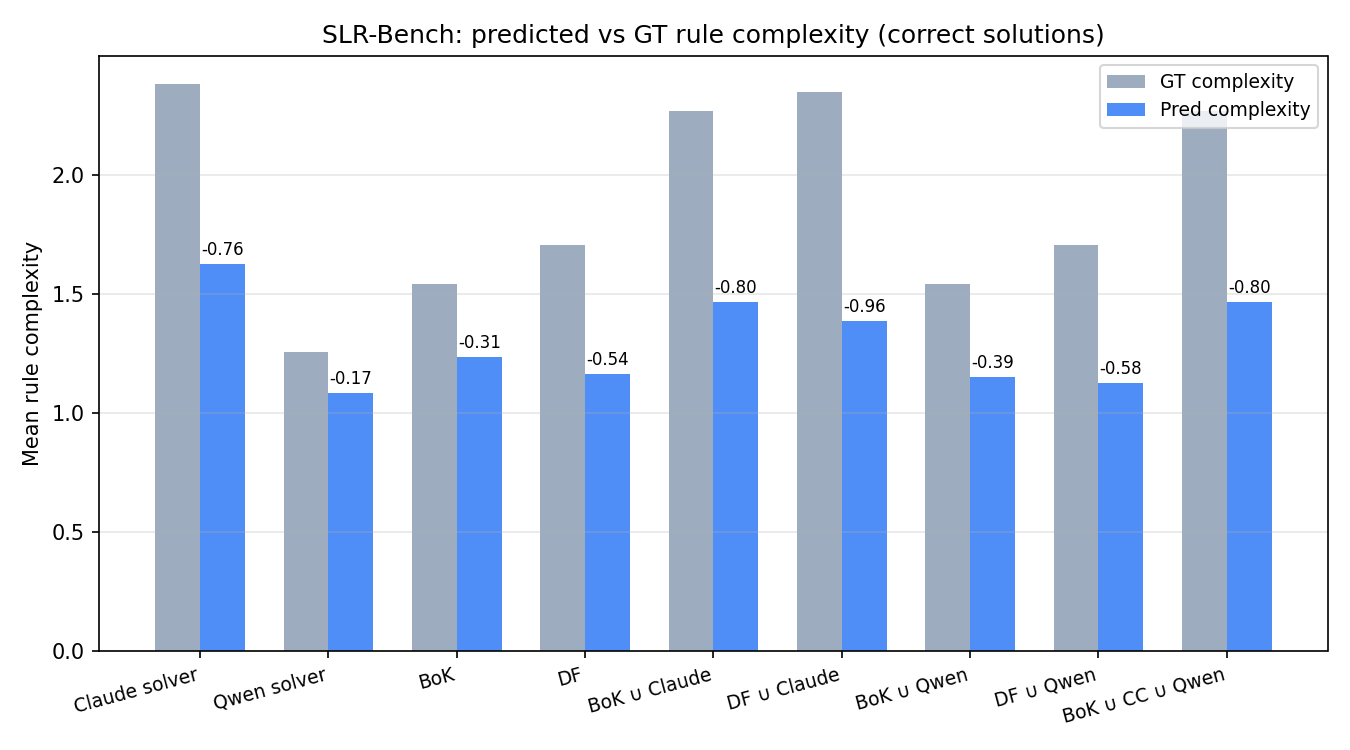}
\caption{\textbf{SLR-Bench: predicted vs.\ ground-truth rule complexity.} All systems predict simpler rules than ground truth; symbolic solvers undershoot most aggressively due to ascending-complexity search.}
\label{fig:slr_complexity}
\end{figure}

\paragraph{Complexity prediction.}
Figure~\ref{fig:slr_complexity} shows predicted vs.\ ground-truth rule complexity for all systems.
In contrast to PBEBench where all systems overshoot ground-truth complexity ($\Delta > 0$), on SLR-Bench all systems predict \emph{simpler} rules than ground truth ($\Delta < 0$), reflecting the abundance of shorter equivalent Prolog forms.
Symbolic solvers undershoot most aggressively ($\Delta \approx -0.76$ to $-1.09$) consistent with their ascending-complexity search that stops at the first correct rule.
LLM-based methods undershoot less ($\Delta \approx -0.61$ to $-0.83$), as they occasionally produce more complex correct rules.

\section{Additional Related Work}
\label{app:more_related_work}
We discuss additional bodies of literature relevant to our work.

\paragraph{Library Learning with LLMs.}
Our method is also related to library learning and abstraction discovery. DreamCoder \citep{ellis2021dreamcoder} is the canonical precursor, jointly learning reusable symbolic abstractions and a neural search policy through wake-sleep library learning. Stitch \citep{bowers2023top} revisits library learning from a symbolic perspective, introducing corpus-guided top-down synthesis for extracting reusable abstractions from a program corpus much more efficiently than prior deductive approaches. LILO \citep{grand2023lilo} builds on this line by combining LLM-guided synthesis with Stitch-style symbolic compression and automatic natural-language documentation of learned abstractions. More recent LLM-based work revisits this idea through tool and abstraction discovery. ReGAL \citep{stengel2024regal} refactors programs into more general reusable abstractions. TroVE \citep{wang2024trove} induces verifiable toolboxes for programmatic tasks. ToolLibGen \citep{yue2025toollibgen} extracts tools from reasoning traces and aggregates them into larger libraries for reuse. Voyager \citep{wang2023voyager} offers a related skill-library perspective in embodied agents, where executable skills are accumulated and retrieved over long horizons. At the same time, a compute-matched re-evaluation of TroVE \citep{sesterhenn2025compute} suggests that some reported gains may shrink once inference-time compute is carefully controlled.

\end{document}